\documentclass[dvipsnames,format=sigconf,anonymous=false,review=false,pbalance=true,nonacm=true,authorversion=true]{acmart}
\usepackage[utf8]{inputenc}
\usepackage{amsmath,mathtools,bbm} % ocio, amssymb messes things with Bbbk
\usepackage[inline,shortlabels]{enumitem}
\usepackage[detect-weight=true]{siunitx}
\PassOptionsToPackage{hyphens}{url}\usepackage{hyperref}
\usepackage{cleveref}
\usepackage{algorithm}
\usepackage[]{algpseudocode}
\crefname{algocf}{alg.}{algs.}
\Crefname{algocf}{Algorithm}{Algorithms}
\usepackage{color,colortbl,etoolbox}
\robustify\bfseries
\usepackage{multirow}
\usepackage{subcaption}
\usepackage{csquotes} %http://tex.stackexchange.com/a/294634
\usepackage[normalem]{ulem}
\usepackage{afterpage}
\usepackage{booktabs}
\renewcommand\vec{\boldsymbol}

\usepackage{stackengine}

\usepackage{glossaries}

\usepackage{tikz}
\usetikzlibrary{external}
\usetikzlibrary{shapes.geometric,fit,backgrounds,decorations.pathreplacing,calligraphy}
\usepackage[linewidth=1pt,ticklabelscale=0.85,titleyshift=-1mm]{plotmacros}

\glsdisablehyper

\usepackage{xspace}
\def\ie{\emph{i.e.}\xspace}
\def\Ie{\emph{I.e.}\xspace}
\def\eg{e.g.\xspace}

\def\vs{vs.\xspace}

\usepackage{svg}

\DeclareRobustCommand{\numcirc}[1]{%
    \tikz[baseline=(N.base), inner sep=0pt]%
    \node(N)[
        circle,
        fill=blue,
        text=white,
        font=\sffamily\bfseries,
        minimum size=1.1em,
        inner sep=1pt
    ] {#1};%
}

\usepackage{listingsutf8}
\lstdefinestyle{llm}{
    basicstyle=\linespread{0.7}\ttfamily\scriptsize\color{darkgray},
    numberstyle=\tiny\color{gray},
    breakatwhitespace=false,
    breaklines=true,
    breakindent=0mm,
    captionpos=b,
    keepspaces=true,
    numbers=left,
    numbersep=5pt,
    showspaces=false,
    showstringspaces=false,
    showtabs=false,
    tabsize=2,
    frame=tlbr,
    framesep=0.5mm,
    framerule=0pt,    
    xleftmargin=\parindent,
    literate={“}{"}1 {”}{"}1 {‘}{`}1 {’}{'}1 {–}{--}1 {…}{...}1 {×}{x}1 {≠}{!=}1 {→}{->}{1}
}
\tikzset{external/only named=true}

\usepackage{lipsum}

\newacronym{ea}{EA}{evolutionary algorithm}
\newacronym{ec}{EC}{evolutionary computation}
\newacronym{ga}{GA}{genetic algorithm}
\newacronym{gp}{GP}{genetic programming}
\newacronym{ge}{GE}{grammatical evolution}
\newacronym{nsga2}{NSGA-II}{non-dominated sorting genetic algorithm 2}
\newacronym{mo}{MO}{multi-objective}
\newacronym{llm}{LLM}{large language model}
\newacronym{cl}{CL}{curriculum learning}
\newacronym{mt}{MT}{multi-task}
\newacronym{ai}{AI}{artificial intelligence}
\newacronym{qd}{QD}{quality diversity}
\newacronym{me}{ME}{MAP-elites}
\newacronym{ann}{ANN}{artificial neural network}

\def\fN{N\xspace}
\def\fNP{N+P\xspace}
\def\fNPB{N+P+B\xspace}
\newcommand{\llm}[1]{{\color{darkgray}\small\texttt{#1}}}

\glsdisablehyper

\copyrightyear{2026}
\acmYear{2026}
\setcopyright{acmlicensed}\acmConference[GECCO '26]{Genetic and Evolutionary Computation Conference}{July 13--17, 2026}{San José, Costa Rica}
\acmBooktitle{Genetic and Evolutionary Computation Conference, July 13--17, 2026, San José, Costa Rica}
\acmDOI{todo}
\acmISBN{todo}

\tikzexternaldisable

\begin{document}

\title[\Acrshort{llm}-assisted \Acrshort{cl} for \Acrshort{mt} evolutionary policy search]{Interactive \acrshort{llm}-assisted Curriculum Learning for Multi-Task Evolutionary Policy Search}

\author{Berfin Sakallioglu}
\orcid{0009-0009-7835-6481}
\affiliation{
    \institution{MIGe - University of Trieste}
    \city{Trieste} 
    \country{Italy} 
}
\email{berfin.sakallioglu@phd.units.it}

\author{Giorgia Nadizar}
\orcid{0000-0002-3535-9748}
\affiliation{
    \institution{University Toulouse Capitole, IRIT - CNRS UMR5505}
    \city{Toulose} 
    \country{France} 
}
\email{giorgia.nadizar@irit.fr}

\author{Eric Medvet}
\orcid{0000-0001-5652-2113}
\affiliation{
    \institution{DIA - University of Trieste}
    \city{Trieste} 
    \country{Italy} 
}
\email{emedvet@units.it}

\begin{abstract} %max 200 words
    \Acrlong{mt} policy search is a challenging problem because policies are required to generalize beyond training cases.
    \Acrlong{cl} has proven to be effective in this setting, as it introduces complexity progressively.
    However, designing effective curricula is labor-intensive and requires extensive domain expertise.
    \Acrshort{llm}-based curriculum generation has only recently emerged as a potential solution, but was limited to operate in static, offline modes without leveraging real-time feedback from the optimizer.
    Here we propose an interactive \acrshort{llm}-assisted framework for online curriculum generation, where the \acrshort{llm} adaptively designs training cases based on real-time feedback from the evolutionary optimization process. 
    We investigate how different feedback modalities, ranging from numeric metrics alone to combinations with plots and behavior visualizations, influence the \acrshort{llm} ability to generate meaningful curricula. 
    Through a 2D robot navigation case study, tackled with \acrlong{gp} as optimizer, we evaluate our approach against static \acrshort{llm}-generated curricula and expert-designed baselines. 
    We show that interactive curriculum generation outperforms static approaches, with multimodal feedback incorporating both progression plots and behavior visualizations yielding performance competitive with expert-designed curricula. 
    This work contributes to understanding how \acrshortpl{llm} can serve as interactive curriculum designers for embodied \acrshort{ai} systems, with potential extensions to broader evolutionary robotics applications.
\end{abstract}

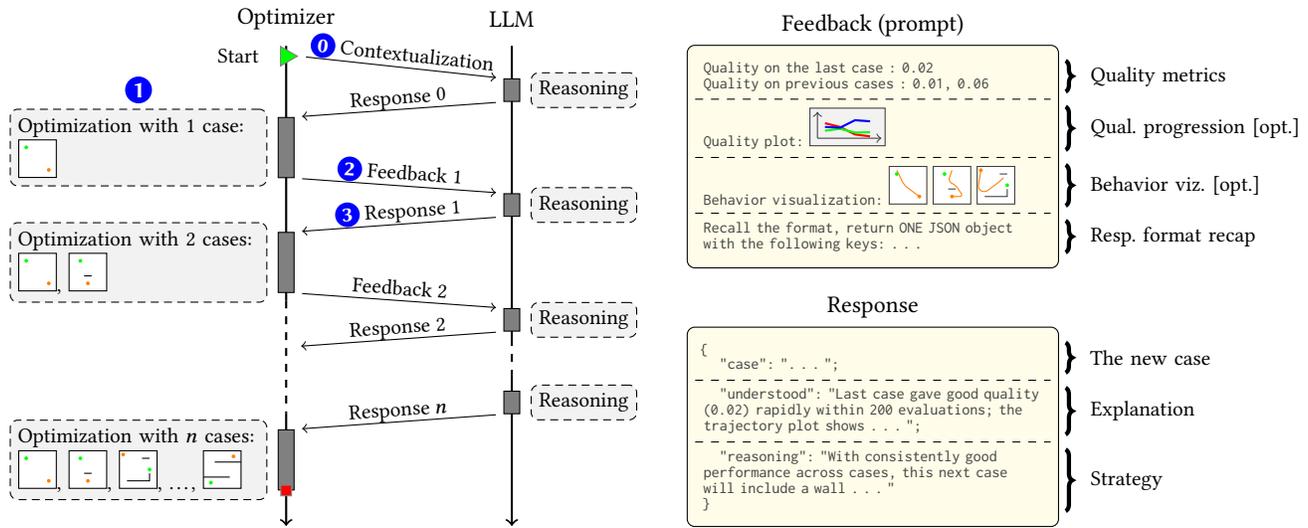
\begin{teaserfigure}
    \centering
    \tikzsetnextfilename{intro-picture}
    \def\bW{2mm}
\def\optBH{1cm}
\def\llmBH{3mm}
\def\msgH{2mm}
\def\gap{3cm}
\pgfsetlayers{background,main}
\tikzset{
    imgs/qplot/.pic = {
        \draw[fill=gray!10] (0,0) rectangle (1,0.5);
        \draw[->] (0.1,0.1) -- (0.95,0.1);
        \draw[->] (0.1,0.1) -- (0.1,0.45);
        \draw[thick,red] (0.2,0.3) -- ++(0.2,-0.05) -- ++(0.2,-0.1) -- ++(0.2,-0.025);
        \draw[thick,green] (0.2,0.2) -- ++(0.2,0.025) -- ++(0.2,-0.05) -- ++(0.2,0.005);
        \draw[thick,blue] (0.2,0.25) -- ++(0.2,-0.01) -- ++(0.2,0.1) -- ++(0.2,-0.015);
    },
    imgs/arena0/.pic = {
        \begin{scope}[solid, thin, rounded corners=0]
            \draw[fill=white] (0,0) rectangle (0.5,0.5);                
            \fill[green] (0.1,0.4) circle (0.025);
            \fill[orange] (0.4,0.1) circle (0.025);
        \end{scope}
    },
    imgs/arena1/.pic = {
        \begin{scope}[solid, thin, rounded corners=0]
            \draw[fill=white] (0,0) rectangle (0.5,0.5);
            \draw (0.2,0.2) -- (0.3,0.2);
            \fill[green] (0.15,0.4) circle (0.025);
            \fill[orange] (0.25,0.1) circle (0.025);
        \end{scope}
    },
    imgs/arena2/.pic = {
        \begin{scope}[solid, thin, rounded corners=0]
            \draw[fill=white] (0,0) rectangle (0.5,0.5);
            \draw (0.1,0.1) -- (0.4,0.1) -- (0.4,0.2);
            \draw (0.3,0.4) -- (0.4,0.4);
            \fill[green] (0.4,0.25) circle (0.025);
            \fill[orange] (0.05,0.45) circle (0.025);
        \end{scope}
    },
    imgs/arena3/.pic = {
        \begin{scope}[solid, thin, rounded corners=0]
            \draw[fill=white] (0,0) rectangle (0.5,0.5);
            \draw (0.0,0.15) -- (0.35,0.15);
            \draw (0.15,0.35) -- (0.5,0.35);
            \fill[green] (0.1,0.075) circle (0.025);
            \fill[orange] (0.4,0.425) circle (0.025);
        \end{scope}
    },
    imgs/behavior0/.pic = {
        \pic {imgs/arena0};                
        \draw[orange, rounded corners=0.5mm] (0.4,0.1) -- (0.2,0.25) -- (0.1,0.45);
    },
    imgs/behavior1/.pic = {
        \pic {imgs/arena1};                
        \draw[orange, rounded corners=0.5mm] (0.25,0.1) -- (0.4,0.1) -- (0.3,0.25) -- (0.2,0.3) -- (0.25,0.45);
    },
    imgs/behavior2/.pic = {
        \pic {imgs/arena2};
        \draw[orange, rounded corners=0.5mm] (0.05,0.45) -- (0.01,0.2) -- (0.1,0.15) -- (0.3,0.3) -- (0.35,0.35);
    }            ,
    imgs/behavior3/.pic = {
        \pic {imgs/arena3};
        \draw[orange, rounded corners=0.5mm] (0.4,0.425) -- (0.1, 0.4) -- (0.1,0.3) -- (0.4,0.2) -- (0.4,0.1) -- (0.2,0.01);
    },
    start/.style={
        regular polygon,
        regular polygon sides=3,
        shape border rotate=-90,
        inner sep=0pt,
        anchor=center,
        minimum width=3mm,
        fill=green,
        draw=gray
    },
    end/.style={
        regular polygon,
        regular polygon sides=4,
        inner sep=0pt,
        anchor=center,
        minimum width=2mm,
        fill=red,
        draw=gray
    }    
}
\begin{tikzpicture}[
    label/.style={
        font=\small,
    },
    desc/.style={
        label,
        align=left,
        rounded corners=1mm,                
        draw=black, densely dashed,
        fill=gray!10
    },
    opt-desc/.style={
        desc,
        anchor=east,
        text width=3.2cm,
        xshift=-0.25cm
    },
    llm-desc/.style={
        desc,
        anchor=west,                
        xshift=0.25cm
    },
    msg/.style={
        ->,
        shorten <=1mm,
        shorten >=1mm
    },
    msg-cnt/.style={
        label,
        midway,
        above=-1mm,
        sloped
    },
    phase/.style={
        draw=black,
        fill=gray,
        anchor=north,
        minimum width=\bW
    },
    opt-phase/.style={
        phase,
        minimum height=8mm
    },
    llm-phase/.style={
        phase,
        minimum height=\llmBH
    }
]
    %optimizer
    \node (opt) at (0,0) {Optimizer};
    \node (o0) at ($(opt)+(0,-0.25)$) {};
    \path (o0) ++(0,-3.5) coordinate (o1) ++(0,-1.5) coordinate (o2) ++(0,-1.5) coordinate (o3);
    \draw[thick] (o0) -- (o1);
    \draw[thick,dashed] (o1) -- (o2);
    \draw[thick,->] (o2) -- (o3);
    %llm
    \node (llm) at (\gap,0) {\acrshort{llm}};            
    \node (l0) at ($(llm)+(0,-0.25)$) {};
    \path (l0) ++(0,-4) coordinate (l1) ++(0,-0.5) coordinate (l2) ++(0,-2) coordinate (l3);
    \draw[thick] (l0) -- (l1);
    \draw[thick,dashed] (l1) -- (l2);
    \draw[thick,->] (l2) -- (l3);
    %phases
    \node[start] (start) at ($(o0)+(0,-0.25)$) {};
    \node[anchor=east,xshift=-0.25cm,label] at (start) {Start};
    \node[llm-phase] (llm0) at ($(start.south -| l0)+(0,-\msgH)$) {};
    \node[llm-desc] at (llm0) {Reasoning};
    \node[opt-phase] (opt1) at ($(llm0.south -| o0)+(0,-\msgH)$) {};
    \node[opt-desc] (opt1desc) at (opt1) {
        Optimization with $1$ case:\\
        \tikz \pic{imgs/arena0};
    };
    \node at ($(opt1desc.north)+(0,0.25)$) {\numcirc{1}};
    \node[llm-phase] (llm1) at ($(opt1.south -| l0)+(0,-\msgH)$) {};
    \node[llm-desc] at (llm1) {Reasoning};
    \node[opt-phase] (opt2) at ($(llm1.south -| o0)+(0,-\msgH)$) {};
    \node[opt-desc] at (opt2) {
        Optimization with $2$ cases:\\
        \tikz \pic{imgs/arena0};,
        \tikz \pic{imgs/arena1};
    };
    \node[llm-phase] (llm2) at ($(opt2.south -| l0)+(0,-\msgH)$) {};
    \node[llm-desc] at (llm2) {Reasoning};
    \node[llm-phase] (llm3) at ($(l2)+(0,-\msgH)$) {};
    \node[llm-desc] at (llm3) {Reasoning};
    \node[opt-phase] (opt3) at ($(llm3.south -| o0)+(0,-\msgH)$) {};
    \node[opt-desc] at (opt3) {
        Optimization with $n$ cases:\\
        \tikz \pic{imgs/arena0};,
        \tikz \pic{imgs/arena1};,
        \tikz \pic{imgs/arena2};,
        \dots,
        \tikz \pic{imgs/arena3};
    };
    \node[end] (end) at (opt3.south) {};
    %msgs
    \draw[msg] (start.east) -- (llm0.north west) node[msg-cnt] {\numcirc{0} Contextualization};
    \draw[msg] (llm0.south west) -- (opt1.north east) node[msg-cnt] {\phantom{p}Response 0\phantom{p}};
    \draw[msg] (opt1.south east) -- (llm1.north west) node[msg-cnt] {\numcirc{2} Feedback 1};
    \draw[msg] (llm1.south west) -- (opt2.north east) node[msg-cnt] {\numcirc{3} Response 1};
    \draw[msg] (opt2.south east) -- (llm2.north west) node[msg-cnt] {\phantom{p}Feedback 2\phantom{p}};
    \draw[msg] (llm2.south west) -- ($(llm2.south -| o0)+(\bW/2,-\msgH)$) node[msg-cnt] {Response 2};
    \draw[msg] (llm3.south west) -- (opt3.north east) node[msg-cnt] {Response $n$};
\end{tikzpicture}
\hspace{0.5cm}
\begin{tikzpicture}[
    prompt/.style={
        font=\linespread{0.8}\ttfamily\scriptsize,
        text=darkgray,
        text width=4.5cm,
        align=left,
        anchor=north
    },
    msg/.style={
        rounded corners=1mm,
        fill=yellow!10,
        draw=black
    },
    separator/.style={
        dashed
    },
    expl-brace/.style={
        decorate,
        decoration={brace},
        ultra thick
    },
    expl-text/.style={
        font=\linespread{0.8}\small,
        midway,
        anchor=west,
        align=left,
        right=2mm,
        text width=3cm
    }
]
    % contextualization
    % \node[prompt] (ctx1) at (0,0) {
    %     You are an assistant for case design in CURRICULUM LEARNING.
    %     I am optimizing the policy of robot for 2D navigation.
    %     \dots
    % };
    % \draw[separator] (ctx1.south west) -- (ctx1.south east);
    % \draw[expl-brace] ($(ctx1.north east)+(0.2,-0.1)$) -- ($(ctx1.south east)+(0.2,0.1)$) node[expl-text] {General info};
    % \node[prompt] (ctx2) at (ctx1.south) {
    %      I will provide some feedback about the previous stage and the progress in general.
    %      Namely, I’ll provide:
    %      \dots
    % };
    % \draw[separator] (ctx2.south west) -- (ctx2.south east);
    % \draw[expl-brace] ($(ctx2.north east)+(0.2,-0.1)$) -- ($(ctx2.south east)+(0.2,0.1)$) node[expl-text] {Feedback format desc.};
    % \node[prompt] (ctx3) at (ctx2.south) {
    %     Return ONE JSON object with the following keys:
    %     \dots
    % };
    % \draw[expl-brace] ($(ctx3.north east)+(0.2,-0.1)$) -- ($(ctx3.south east)+(0.2,0.1)$) node[expl-text] {Resp.\ format prescription};
    % \begin{scope}[on background layer]
    %     \node[msg,label=above:Contextualization (prompt),fit=(ctx1) (ctx2) (ctx3)] {};
    % \end{scope}
    % feedback
    \node[prompt] (feedback1) at (0,0) {
        Quality on the last case : 0.02\\
        Quality on previous cases : {0.01, 0.06}
    };
    \draw[separator] (feedback1.south west) -- (feedback1.south east);
    \draw[expl-brace] ($(feedback1.north east)+(0.2,-0.1)$) -- ($(feedback1.south east)+(0.2,0.1)$) node[expl-text] {Quality metrics};
    \node[prompt] (feedback2) at (feedback1.south) {
        Quality plot:
        \tikz \pic{imgs/qplot};
    };
    \draw[separator] (feedback2.south west) -- (feedback2.south east);
    \draw[expl-brace] ($(feedback2.north east)+(0.2,-0.1)$) -- ($(feedback2.south east)+(0.2,0.1)$) node[expl-text] {Qual.\ progression [opt.]};
    \node[prompt] (feedback3) at (feedback2.south) {
        Behavior visualization:
        \tikz \pic{imgs/behavior0};
        \tikz \pic{imgs/behavior1};
        \tikz \pic{imgs/behavior2};
    };
    \draw[separator] (feedback3.south west) -- (feedback3.south east);
    \draw[expl-brace] ($(feedback3.north east)+(0.2,-0.1)$) -- ($(feedback3.south east)+(0.2,0.1)$) node[expl-text] {Behavior viz.\ [opt.]};
    \node[prompt] (feedback4) at (feedback3.south) {
        Recall the format, return ONE JSON object with the following keys: \dots
    };
    \draw[expl-brace] ($(feedback4.north east)+(0.2,-0.1)$) -- ($(feedback4.south east)+(0.2,0.1)$) node[expl-text] {Resp.\ format recap};
    \begin{scope}[on background layer]
        \node[msg,label=above:Feedback (prompt),fit=(feedback1) (feedback2) (feedback3) (feedback4)] {};
    \end{scope}
    % response
    \node[prompt] (resp1) at ($(feedback4.south)+(0,-1)$) {
        \{ \\
        \phantom{aa}"case": "\dots";
    };
    \draw[separator] (resp1.south west) -- (resp1.south east);
    \draw[expl-brace] ($(resp1.north east)+(0.2,-0.1)$) -- ($(resp1.south east)+(0.2,0.1)$) node[expl-text] {The new case};
    \node[prompt] (resp2) at (resp1.south) {
        \phantom{aa}"understood": "Last case gave good quality (0.02) rapidly within 200 evaluations; the trajectory plot shows \dots";
    };
    \draw[separator] (resp2.south west) -- (resp2.south east);
    \draw[expl-brace] ($(resp2.north east)+(0.2,-0.1)$) -- ($(resp2.south east)+(0.2,0.1)$) node[expl-text] {Explanation};
    \node[prompt] (resp3) at (resp2.south) {
        \phantom{aa}"reasoning": "With consistently good performance across cases, this next case will include a wall \dots"\\
        \}
    };
    \draw[expl-brace] ($(resp3.north east)+(0.2,-0.1)$) -- ($(resp3.south east)+(0.2,0.1)$) node[expl-text] {Strategy};
    \begin{scope}[on background layer]
        \node[msg,label=above:Response,fit=(resp1) (resp2) (resp3)] {};
    \end{scope}
\end{tikzpicture}

    \caption{
        An overview of our fully unsupervised pipeline for performing \gls{cl} policy search using a curriculum generated by an \acrshort{llm} based on the feedback on the optimization process.
        At the beginning (\protect\tikz[baseline={([yshift=-0.5ex]current bounding box.center)}] \protect\node[start] {};) the optimizer \protect\numcirc{0} sends a contextualization prompt to the \acrshort{llm} describing the general context, the subsequent interactions, and the format of feedback and response messages, and asking for an initial training case.
        Then, the optimizer repeats the following steps, starting with a set comprising the initial \acrshort{llm}-provided training case: \protect\numcirc{1} it performs an optimization run, \protect\numcirc{2} it sends a summary of this run to the \acrshort{llm}, asking for a new case, \protect\numcirc{3} it gets a response with a new case and adds it to the training set.
        After $n$ iterations of \protect\numcirc{1}\protect\numcirc{2}\protect\numcirc{3}, the pipeline stops (\protect\tikz[baseline={([yshift=-0.5ex]current bounding box.center)}] \protect\node[end] {};) and the optimizer returns the best policy found in the process.
        On the right, we sketch the structure of the feedback prompt and the \acrshort{llm} response, highlighting the salient components.        
    }
    \label{fig:intro-picture}
\end{teaserfigure}




\maketitle
\glsresetall

\section{Introduction and related works}
\label{sec:intro}
\Gls{mt} learning trains a single model to solve several related tasks jointly instead of in isolation, encouraging shared representations that improve generalization to new but similar tasks~\cite{caruana1997multitask,zhang2021survey}.
In robotics, for example, a visuomotor policy might be trained across families of manipulation tasks such as pushing objects with different shapes, opening doors and drawers with varying kinematics, or placing items into containers at different poses; these tasks share perception and embodiment but differ in contact dynamics and success criteria~\cite{devin2017learning,yang2020multi}. 
Learning them in an \gls{mt} setting is particularly important for real-world deployment, where agents might be called to rapidly adapt to novel tasks that are close variants of those seen during training, rather than repeatedly relearning from scratch~\cite{hessel2019multi}.

Evolutionary policy search provides a general framework for optimizing \gls{mt} policies without relying on gradients.
By iteratively selecting, mutating, and recombining candidate solutions, evolutionary policy search can optimize arbitrary policy representations, including interpretable~\cite{nadizar2024naturally,de2025evolution} or structured policies discovered via \gls{gp}~\cite{vacher2025maple}. 
This flexibility makes evolutionary optimization particularly interesting for \gls{mt} settings, where diverse~\cite{mouret2020quality,nadizar2024searching} or structured~\cite{desnos2021gegelati} solutions may be needed.
For example, \gls{gp} has been applied in \gls{mt} visual learning domains~\cite{jaskowski2008multitask}---such as evolving controllers that perform well across multiple Atari games or diverse image processing tasks~\cite{bi2021learning,bi2022multitask}---demonstrating that symbolic policies can capture shared structure while remaining potentially human-interpretable.

The iterative nature of evolutionary policy search naturally interacts with \gls{cl}~\cite{jorgensen2024large,jorgensen2025policy,liu2025curriculum}: by controlling how and when tasks (or data from each task) are presented, curricula can guide populations toward solutions that capture shared structure more efficiently~\cite{bengio2009curriculum}.
In \gls{mt} settings, curricula often stage tasks so that simpler or information-rich tasks bootstrap learning of features or structures that later support harder problems~\cite{pentina2015curriculum}.

However, designing effective curricula poses a fundamental challenge. Crafting representative, difficulty-calibrated training instances or auxiliary tasks typically requires substantial domain expertise~\cite{bengio2009curriculum,soviany2022curriculum}. 
In robotics, this often entails manually creating synthetic trajectories or intermediate challenges to target specific capability gaps, frequently without access to real-world data~\cite{florensa2017reverse}. 
Moreover, even when suitable tasks are available, scheduling them effectively is nontrivial~\cite{matiisen2019teacher}.

Adaptive curricula offer a promising alternative by adjusting task sampling online based on learning progress, validation performance, or model uncertainty, focusing training on tasks that are neither too easy nor too difficult~\cite{graves2017automated,varshney2022let}.
Such approaches can improve transfer, aligning training pressure with the agent evolving capabilities to enhance generalization~\cite{jean2019adaptive,matiisen2019teacher,xu2022statistical,wang2025gacl}. 
Yet jointly designing task content and scheduling---especially when done by humans---adds significant complexity, requiring careful supervision and continual intervention during training~\cite{soviany2022curriculum}.

In this context, \glspl{llm} provide a promising path toward fully automated curriculum design. 
They can generate diverse tasks, environments, reward functions, or curricula, either statically~\cite{ryu2025curricullm,ma2023eureka} or adaptively~\cite{faldor2024omni,liang2024environment,li2025dynamic}. 
\Glspl{llm} have been used to propose initial environments for training~\cite{liang2024environment}, dynamically adjust difficulty based on agent performance~\cite{li2025dynamic}, and even produce structured teaching curricula in educational settings~\cite{zhang2025eduplanner}. 
Interactive \gls{llm}-based \gls{cl} approaches, such as EnvGen~\cite{zala2024envgen}, combine task generation with feedback from the agent, allowing the curriculum to progress alongside the learner. 
% By producing rich, diverse training scenarios and estimating task difficulty, \glspl{llm} can guide \gls{mt} learning efficiently without manual intervention.

Building on these advances, we propose interactive \gls{llm}-assisted curriculum learning for \gls{mt} evolutionary policy search. 
Our approach uses an \gls{llm} to generate candidate tasks or task variants, receive feedback from evolutionary optimization signals, and iteratively refine the curriculum to match the population evolving capabilities. 
Unlike prior methods~\cite{zala2024envgen}, we never train agents on hand-designed test cases.
In addition, our method supports potentially interpretable controller representations and explores different feedback modalities, including multimodal signals, which can further improve learning~\cite{huang2025multimodal}.

We test our approach on a 2D navigation task using a symbolic equation-based controller and find that interactive \gls{llm}-generated curricula outperform static \gls{llm}-generated curricula, with richer feedback further boosting performance. 
Overall, our method achieves results comparable to expert-designed curricula with zero human effort, demonstrating that it can efficiently guide \gls{mt} learning, even for interpretable representations. 
Moreover, this also opens the door to larger-scale applications, as \glspl{llm} can inherently generate a wide variety of environments~\cite{faldor2024omni} and can even be fine-tuned to produce specialized ones~\cite{sudhakaran2023mariogpt,aki2024llm}.

\section{Interactive \acrshort{llm}-assisted \acrlong{cl}}
\label{sec:general-working}
We propose a fully unsupervised pipeline for performing \gls{cl} policy search using a curriculum generated by an \gls{llm} based on the feedback concerning the optimization process.
The pipeline, sketched in \Cref{fig:intro-picture}, involves two actors, the \emph{optimizer} and the \emph{\gls{llm}}, interacting iteratively through messages for a given number of times.

We assume that the problem being tackled can be instantiated in \emph{cases}, \ie, problem instances on which a candidate policy can be evaluated.
Moreover, we assume that the optimizer can perform the optimization on a bag of one or more cases, called training cases.

The optimizer-\gls{llm} interaction starts with a \numcirc{0} \emph{contextualization message} sent by the optimizer to the \gls{llm} to provide basic information and context about the overall interaction and to ask for an initial training case $c_0$.
This message serves for instructing the \gls{llm} about the subsequent interactions, to describe the format of feedback messages, and to prescript the format of responses, namely on how to encode new cases.

After this first exchange, the interaction works as follows, starting from a bag $C=\{c_0\}$ containing the single initial training case provided by the the \gls{llm}.
For a given number $n_\text{stage}$ of times,
\begin{enumerate*}[label={}]
    \item[\numcirc{1}] the optimizer runs an optimization run on $C$,
    \item[\numcirc{2}] the optimizer sends a \emph{feedback message} to the \gls{llm} containing the feedback about the optimization run,
    \item[\numcirc{3}] the \gls{llm} responds with a message containing a new case $c$, then the bag of training cases is augmented with $c$, $C \gets C \oplus \{c\}$.
\end{enumerate*}
We call \emph{stage} an iteration of the three steps above.
We call \emph{curriculum} the sequence of training bags obtained by the optimizer at the end of the process.

The precise format of messages (\ie, the corresponding \emph{prompts}) sent by the optimizer is problem-specific.
We describe them in detail later for the case study which we consider for assessing our proposal.
\Cref{fig:intro-picture} (on the right) sketches the structure of these messages, highlighting the main components; Appendix~\ref{sec:appendix-example} shows them in detail.

We remark that, at each stage, the bag $C$ includes the cases of all the previous stages: we opt for accumulating training cases, rather than training at each stage on a single case, for avoiding catastrophic forgetting, \ie, having the optimizer producing a policy that solves the current case but ``forgets'' how to solve the previous ones, eventually hindering generalization~\cite{french1999catastrophic}.
As an aside, this relives the user from the burden of choosing an optimizer and a policy representation which are robust to catastrophic forgetting.

\subsection{Feedback modality}
\label{sec:feedback-modalities}
We investigate three feedback modalities that progressively increase information richness:
\begin{description}
    \item[Numeric feedback (\fN)] contains only quantitative performance metrics, namely, the quality scores of the policy found by the optimizer assessed on each case in $C$.
    \item[Progression-augmented feedback (\fNP)] extends \fN by adding a visualization of optimization convergence. 
    This plot shows the quality of the best policy across optimization iterations ($x$-axis: iteration number; $y$-axis: quality metric), one line for each case in $C$, revealing how quickly and effectively the optimization process improved.
    \item[Behavioral-augmented feedback (\fNPB)] further extends \fNP by including a visualization of the policy behavior on the training cases.
    This provides the \gls{llm} with direct insight into the policy behavioral characteristics and how it interacts with the problem constraints.
    In the case study considered here (see \Cref{sec:case-study}), behavior is visualized as a spatial trajectory showing the path taken by the agent overlaid on the environment layout.
\end{description}

\subsection{Requirements and limitations}
Our pipeline is general and agnostic with respect to the inner working of the optimizer, the nature of the policy, and the considered problem.
Nevertheless, there are some requirements concerning the optimizer and the problem.

The optimizer must:
\begin{enumerate*}[(a)]
    \item accept multiple training cases that accumulate over stages,
    \item produce a numeric quality measure quantifying the policy performance on each case in $C$, and
    \item if \fNP or \fNPB feedback is employed, track and report the quality progression across optimization iterations for each case.
\end{enumerate*}
Moreover, it is convenient, but not strictly required, that the optimizer can start an optimization from the results of a previous optimization run: this makes the entire process more efficient and effective.

The problem must:
\begin{enumerate*}[(a)]
    \item allow for a case representation which can be encoded in a format the \gls{llm} can generate and
    \item if \fNPB feedback is employed, allow for a visual representation of the agent behavior on a case, given a policy.
\end{enumerate*}
The first problem requirement might appear hard to meet, as the syntax describing a case might be custom and not known to the \gls{llm}.
However, current \glspl{llm} are becoming increasingly better at complying with provided guidelines~\cite{sun2024ai,wang2025beyond}.
And there are also some \glspl{llm} which have been customized for specific domains, \eg, MarioGPT~\cite{sudhakaran2023mariogpt} used for generating levels of the Super Mario Bros game in \cite{jorgensen2025policy}.
Moreover, in many cases one can develop a procedure that automatically fixes small issues in \gls{llm}-generated cases.
In the problem considered as case study in this work, the \gls{llm} was in general able to generate syntactically correct cases.

\section{Case study: 2D navigation}
\label{sec:case-study}
We consider as use case the problem of learning a policy that permits a simulated robotic agent to navigate \emph{any} 2D arena avoiding obstacles and reaching a target which can perceive by the agent through sensors.

\subsection{Model, policy, and optimization}

\subsubsection{Agent and environment}
The agent is a simulated differential-drive robot with two wheels, five proximity sensors, and two sensors for sensing the distance to and the relative direction of the target.
The environment is a 2D arena defined by a robot starting position, a target position, and some obstacles in the form of walls placed inside the arena.
The size of the arena is of $\qty{1}{\meter} \times \qty{1}{\meter}$; the robot has a circular shape with a radius of \qty{0.02}{\meter}.

We simulate the movement of the robot in the arena in discrete time with a time step of \qty{0.1}{\second}.
At each time step, the robot perceives the distance to the closest obstacle (arena internal and external walls) along five directions equally spread on the front side of the robot (\ie, $-\frac{\pi}{2}$, $-\frac{\pi}{4}$, $0$, $\frac{\pi}{4}$, and $\frac{\pi}{2}$), the distance to the target, and the relative direction to the target.
The proximity sensors and the distance-to-target sensor have a range of \qty{0.5}{\meter}: if there are no objects in the range, the sensors sense \qty{0.5}{\meter}; the domain of these sensors readings is hence $[0,0.5]$.
The angle-to-target sensor domain is $[-\pi,\pi]$.

Upon the processing of the sensory input, the robot sets independently the rotational speed of the two wheels in $[-\omega_\text{max},\omega_\text{max}]$.
We set $\omega_\text{max}$ such that the maximum linear speed of the robot is \qty{0.01}{\meter\per\second}.
When the robot collides with an obstacle, the linear speed goes to zero but we allow it to rotate.

\subsubsection{Policy representation}
Given the sensors and actuators of the robot, its controller is in general a dynamical system with $\mathbb{R}^7$ as observation space and $\mathbb{R}^2$ as action space.
We implement the controller as an outer fixed part wrapping an inner variable part, which is the one we actually optimize.

The outer part pre-processes the observation for the wrapped inner part and post-processes the latter output before using it as actuation values for the wheels. 
In the pre-processing, it takes the original observation $\vec{o}^{(k)} \in \mathbb{R}^7$ and
\begin{enumerate*}[(i)]
    \item normalizes each element in the proper domain (\ie, $[0,0.5]$ or $[-\pi,\pi]$) and
    \item augments it with trend and average value during the last \qty{0.5}{\second}: \ie, for each $o_i^{(k)}$, the augmented observation $\vec{o}'^{(k)} \in [-1,1]^{21}$ contains normalized $o_i^{(k)}$, $o_i^{(k)}-o_i^{(k-4)}$, and $\frac{1}{5}\sum_{j=0}^{j=4} o_i^{(k-j)}$.
\end{enumerate*}
In the post-processing, the outer part takes the output $a' \in \mathbb{R}$ of the inner part and translates it to $\vec{a}^{(k)} \in [-\omega_\text{max},\omega_\text{max}]$, with $a_1^{(k)}=a_\text{left}^{(k)}=\omega_\text{max} \left(1+2 \min \left(0, \tanh a'^{(k)}\right)\right)$ and $a_2^{(k)}=a_\text{right}^{(k)}=\omega_\text{max} \left(1-2 \max \left(0, \tanh a'^{(k)}\right)\right)$.
This way, if $a'^{(k)} \to -\infty$ then $\vec{a}^{(k)}=(-\omega_\text{max},\omega_\text{max})$ and the robot rotates to the left, if $a'^{(k)}\to +\infty$ then $\vec{a}^{(k)}=(\omega_\text{max},+\omega_\text{max})$ and it rotates to the right, if $a'^{(k)}=0$ then $\vec{a}^{(k)}=(\omega_\text{max},\omega_\text{max})$ and it goes straight at the maximum speed.
In the intermediate cases, the robot follows a curvy path, with the radius of the curve depending on $|a'^{(k)}|$.
Note that the outer part of the controller is stateful in the pre-processing, as it maintains a memory of the last $5$ observations, and stateless in the post-processing.

As inner part of the controller we use a function in $\mathbb{R}^{21} \to \mathbb{R}$ which we encode through a syntax regression tree, suitable to be optimized with \gls{gp}.
Namely, we use the following \num{11} labels for the non-terminal nodes: $\bullet+\bullet$, $\bullet-\bullet$, $\bullet \times \bullet$, $\bullet \div^* \bullet$, $\log^*\bullet$, $\max(\bullet,\bullet)$, $\min(\bullet,\bullet)$, $\tanh \bullet$, $\bullet>\bullet$, $\bullet<\bullet$, $\bullet?\bullet:\bullet$, where bullets represent the children of the corresponding node and hence dictate its arity.
$\div^*$ and $\log^*$ represent the protected versions of the corresponding operators.
The comparison operators $>$ and $<$ return the sign of the difference of the two arguments; $\bullet?\bullet:\bullet$ corresponds to the ternary operator where the first argument is compared against $0$.
For the terminal nodes we use as labels the observation variables $o'^{(k)}_1,\dots,o'^{(k)}_{21}$ and ephemeral constants in $[-5,5]$.

Summarizing, we drive the robotic agent with a \gls{gp} tree encoding a function from $\mathbb{R}^{21}$ to $\mathbb{R}$ wrapped within a dynamical system which augments sensor readings with ``historical'' information and simplifies the action space.
We call policy the inner tree.

\subsubsection{Tasks}
The overall goal is to find a policy allowing a robot to navigate any arena, \ie, able to solve the general problem of reaching a target while avoiding collisions.
We attempt to achieve this goal by making candidate policies experience different arenas, each one constituting a \emph{training case}.
We call episode a simulation of a robot with a policy within an arena.
In our settings, an episode is deterministic.

Given an arena with a starting position for the robot and a position for the target, we run an episode for \qty{60}{\second} (simulated time, \ie, for \num{600} time steps).
At the beginning of the episode, we set the direction of the robot to the right, regardless of the position of the target.
At the end of episode, we record the full trajectory of the robot.
Out of this, we compute several metrics, including the final (\ie, at $k=600$) and average distance of the robot to the target.

We remark that this 2D navigation problem has been widely used in previous work as a basic, yet not trivial, benchmark for policy search, often for assessing \gls{qd} search algorithms~\cite{grillotti2023kheperax,ludwig2025trace,bahlous2025dominated,medvet2025quality}.
However, in most of the cases, policies were assessed only on the arena they were trained on.
Instead, here we aim at obtaining a policy able to navigate, in general, any arena, which is a much harder problem.
\Ie, we cast this navigation problem as a \emph{\acrfull{mt}} problem where a single policy is required to solve many (similar) tasks, each one defined by an arena.

\begin{figure}
    \centering
    \includegraphics[width=1\linewidth]{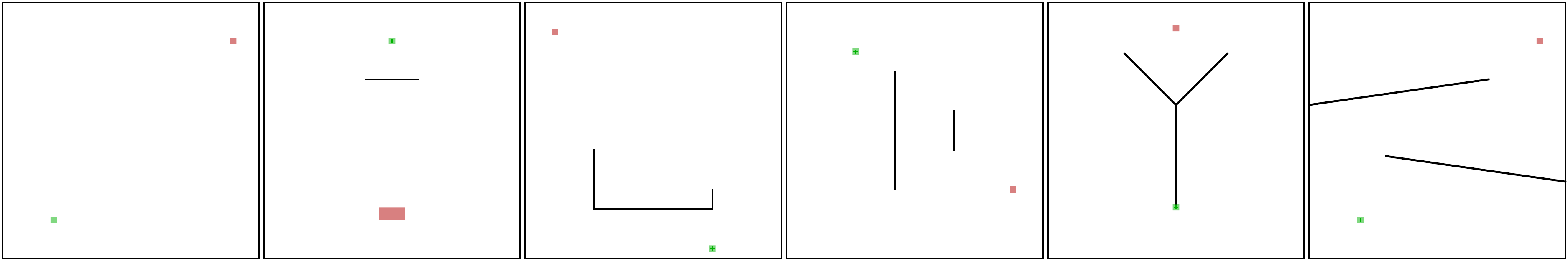}    
    \caption{
        The six test arenas used for assessing policies.
        The red region shows the robot starting position (which is placed at that center of the region), the green region shows the target position.
    }
    \label{fig:test-arenas}
\end{figure}

To assess the effectiveness of a policy to solve the \gls{mt} navigation problem, we consider the set $C_\text{test}$ of six test arenas of \Cref{fig:test-arenas}.
We remark that these six test arenas are never known to the optimizer, nor to the \gls{llm}.

\subsubsection{Evolutionary optimizer}
Based on the recent literature involving the 2D navigation problem, we use \gls{me} as optimizer.
``Classic'' \gls{me}~\cite{mouret2015illuminating} evolves a population of candidate solutions using only mutation as variation operator.
Solutions compete to occupy a cell in a discrete structure called archive: \gls{me} maps solutions to cells using two descriptors which provide numerical coordinates which are then discretized.
Here we use a grid-based archive, where discretization corresponds to matching continuous values of each descriptor against equally sized bins in the corresponding domain.

Given a search space $S$ (here, the set of \gls{gp} trees used as policies), a fitness function $f: S \to \mathbb{R}$, and two descriptors $d_1: S \to D_1 \subset \mathbb{R}$ and $d_2: S \to D_2 \subset \mathbb{R}$, our \gls{me} works as follows---we assume that $f$ has to be minimized.

First, we initialize a population of $n_\text{pop}$ candidate solutions and insert each solution in the archive.
To this end, for each solution $s$, we compute its coordinates $\vec{d}=\left(d_1(s)\vert_{D_1,n_\text{bins}},d_2(s)\vert_{D_2,n_\text{bins}}\right)$ in the archive, where $x\vert_{X,n}=\left\lfloor n \frac{x-\min_{x' \in X}}{\max_{x' \in X} - \min_{x' \in X}}\right\rfloor \in \{0,\dots,n\}$ denotes the discretization of $x$ using $n$ equally sized bins in $X \subset \mathbb{R}$.
If the cell is empty, we put $s$ in the archive.
If the cell contains a solution $s'$, we replace it with $s$ only if $f(s) \le f(s')$.
After the initialization, we repeat the following steps until some termination criterion is met.
\begin{enumerate*}[(i)]
    \item We generate an offspring of $n_\text{batch}$ new solutions by extracting, randomly with repetitions, $n_\text{batch}$ solutions from the archive and mutating each one of them.
    \item We possibly insert each new individual in the archive, as for the initial population.
\end{enumerate*}
At the end of the evolution, the archive contains a set of solutions, \ie, policies, which are diverse based on the descriptors and are good based on $f$.

For initializing the population at the first stage, we use the ramped-half-and-half procedure, commonly used in \gls{gp}, with tree depth enforced to be in $[4, 10]$.
For initializing the population at subsequent stages, we take the best $\frac{1}{4}n_\text{pop}$ solutions of the last iteration population of the previous stage, one mutated solution obtained from each of them, and generate the remaining $\frac{1}{2}n_\text{pop}$ solutions randomly with the ramped-half-and-half procedure: this way, we seed the initial population of the $i$-th stage with the outcome of the optimization of the $i-1$-th stage.

As mutation, we use with equal probability the \gls{gp} subtree mutation or a Gaussian perturbation of the terminal nodes labeled with numerical constants, with a $\sigma=0.25$: when mutating trees, we enforce a limit on the tree depth $\le 10$.

\paragraph{Fitness function and descriptors}
Given a bag $C$ of training cases (here, arenas) and a candidate solution $s$ (here, a tree), we compute the fitness as the weighted average of the final and average distance of the robot to the target, averaged across arenas, \ie, $f(s;C)=\frac{1}{|C|} \sum_{c \in C} f_\text{single}(s;c)$, where:
\begin{equation*}
    f_\text{single}(s;c)=0.9 \left\lVert \vec{x}_\text{robot}^{(600)}-\vec{x}_\text{target} \right\rVert+0.1 \frac{1}{600} \sum_{k=1}^{k=600}\left\lVert \vec{x}_\text{robot}^{(k)}-\vec{x}_\text{target} \right\rVert
\end{equation*}
is the weighted average on a single arena $c$, with $\vec{x}^{(k)}_\text{robot}$ being the position of the robot equipped with $s$ at step $k$ of an episode in the arena $c$ and $\vec{x}_\text{target}$ being the target position in $c$.
We choose to use an average of the final and average distance to the target, and not just the final distance, for driving the optimization, as this leads to ``smarter'' behaviors, awarding policies resulting in shorter trajectories to the target.
However, we assign a relatively low weight ($0.1$) to the average distance as we verified that larger values make the search harder, as certain arenas may require to increase the length of the path (and hence worsen the average distance) to overcome a local optimum.

As \gls{me} descriptors, we considered the gap along the $x$ and $y$ axes of the robot to the target at the end of the episode, averaged across arenas, \ie, $d_1(s;C)=\frac{1}{|C|}\sum_{c \in C} \left(x_{1,\text{robot}}^{(600)}-x_{1,\text{target}}\right)$ and $d_2(s;C)=\frac{1}{|C|}\sum_{c \in C} \left(x_{2,\text{robot}}^{(600)}-x_{2,\text{target}}\right)$.
These descriptors correspond to the adaptation of the classic \gls{me} descriptors used for the 2D navigation problem on a single arena to the case where a policy is assessed on many arenas.
In the classic case, the final coordinates of the robot are used as descriptors: this would not make sense here as different arenas have in general different target position, obstacles, and starting position.
We hence consider the relative position to the target and average it across arenas.
We set the descriptor domains to $D_1=D_2=[-0.5,0.5]$.

\paragraph{Parameters and remark}
Based on experience, previous work, and exploratory experiments, we set the parameters of the optimizer as follows: $n_\text{pop}=n_\text{batch}=100$, $n_\text{bin}=10$, and, as termination criterion, the number of fitness evaluations being performed (which we set to \num{10000} or more depending on the experiment).
We recall that a single fitness evaluation corresponds to performing $|C|$ episodes.

We verified experimentally that this combination of policy representation and optimizer is effective enough to tackle the \gls{mt} 2D navigation problem.
In particular, we compared it against combinations of \gls{ga} as optimizer, \glspl{ann} as policies, and different \gls{me} descriptors based on both the behavior or on policy features: we found that \gls{me} with the descriptors based on the relative final position and \gls{gp} trees for representing the policy is the best combination.
We remark, however, that the main contribution of this work is the optimizer-\gls{llm} interaction scheme, which is by design agnostic to these lower level details.

\subsection{\acrshort{llm} interaction in the 2D navigation case}
The overall interaction scheme described in Section~\ref{sec:general-working} is general, and many components of the prompt remain domain-agnostic. 
However, certain elements are problem-dependent and require customization for the 2D navigation task. 
Here we describe these problem-specific adaptations.

As shown in \Cref{fig:intro-picture}, we begin with a contextualization prompt that differs from subsequent feedback prompts. 
This initial prompt establishes the 2D navigation domain, defining the differential drive robot task where the agent must navigate from a start position to a target while avoiding walls. 
The prompt specifies the role of the \gls{llm} as a stage designer, defines the quality metric (distance to target), describes the feedback format it will receive, enforces validity constraints, and provides an example arena.
We give the complete contextualization prompt in Appendix~\ref{sec:appendix-example}.

We communicate arena configurations through character string encodings.
We assume that an arena is described by tiles and each tile is encoded as a character: \llm{s} represents the start position, \llm{t} the target, \llm{w} denotes walls, and \llm{e} indicates empty tiles. 
The \gls{llm} generates $15\times15$ character grids with rows separated by the \llm{|} character. 
A substantial portion of the prompt enforces structural constraints, including exactly one start and one target position, guaranteed reachability via a continuous path, consistent row lengths of $15$ characters, and internal validation of connectedness. 
The complete format specification is visible in the prompt shown in Appendix~\ref{sec:appendix-example}..

Subsequent feedback prompts vary by modality. 
For \fN, we provide quality scores including the best policy distance to target on the current case, historical performance on previous cases, and average quality across all accumulated cases. 
\fNP extends this with a convergence plot (PNG format) showing distance to target ($y$-axis) over fitness evaluations ($x$-axis), with separate lines for each accumulated case to reveal optimization dynamics.
A legend identifies each case line in the plot. 
\fNPB further adds trajectory visualizations---2D plots (SVG format) similar to \Cref{fig:expert-arenas} displaying walls, start position, target, and the spatial path taken by the best-performing policy at the end of training for each case. 
A feedback example showing the convergence plot and the trajectory visualization is shown in \Cref{sec:llm-reasoning-insights}; a complete interaction sequence is presented in Appendix~\ref{sec:appendix-example}.

\section{Experimental Evaluation}
\label{sec:experiments}
We evaluate the proposed pipeline through a series of experiments comparing different feedback modalities and baseline approaches.
Our experimental design addresses the following research questions:
\begin{enumerate}[label=RQ\arabic*]
    \item \label{item:rq1-general-effectiveness}Does the pipeline enable learning, with policies improving across curriculum stages and generalizing to unseen cases?
    How do \gls{llm}-generated curricula compare to expert-designed, random, and static baselines?
    Which feedback modality is the most effective among the three investigated conditions?
    \item \label{item:rq2-progressiveness}Are \gls{llm}-generated curricula actual curricula?
    Do their cases work equally well if administrated in a single batch instead of progressively?
    %todo
    \item \label{item:rq3-reasoning}Does the \gls{llm} really ``understand'' the feedback?
    Does it exploit it when creating new cases?
\end{enumerate}

\subsection{Experimental procedure and baselines}
\begin{figure}
    \centering
    \includegraphics[width=1\linewidth]{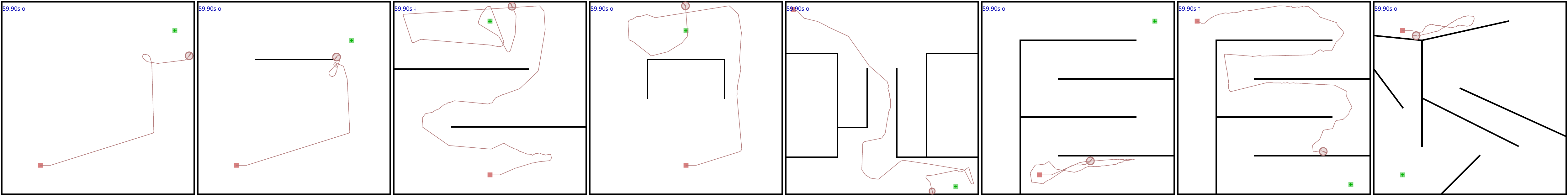}    
    \caption{
        The expert-designed curriculum with the trajectories obtained with a policy optimized on this curriculum.
    }
    \label{fig:expert-arenas}
\end{figure}

We compare three versions of our pipeline corresponding to the feedback modalities described in \Cref{sec:feedback-modalities} against three baselines without feedback.
\begin{description}
    \item[Expert] Represents an expert-designed static curriculum of eight cases (see \Cref{fig:expert-arenas}) created by a domain expert (one of the authors of this paper), serving as our gold standard.
    We generated this curriculum based on our (quite long) experience on the navigation problem and evolutionary approaches for solving it; we also knew the test arenas.    
    \item[Static] Consists of an \gls{llm}-generated curriculum which we obtain by asking the \gls{llm} to build eight cases in a single batch without iterative feedback, using a modified prompt.
    \item[Random] Comprises randomly generated cases with randomized wall segments of varying lengths, numbers, and orientations, as well as random start and target positions.
\end{description}
All curricula have the same number of cases (eight).
We administer all of them progressively, in eight stages: first stage with the first arena, second with the first and the second arenas, and so on.

We repeat the execution of each method $10$ times to assess statistical significance and consistency.
For the variants (both with and without feedback) involving the \gls{llm}, we execute the entire process $10$ times, resetting the \gls{llm} session at each repetition.
For the random baseline, we generate $10$ random curricula.
For the expert baseline, we execute $10$ sequences of optimization runs with different random seeds on the same expert-designed curriculum.

We use JGEA~\cite{medvet2022jgea} for performing the optimization with \gls{me} and for simulating the navigation episodes.
We use Claude Sonnet 4.5 \cite{anthropic2025sonnet45} with extended thinking capabilities as the \gls{llm}---we leave to future work an investigation on the impact of using other \glspl{llm}.
We forwarded the messages between the optimizer (JGEA) and the \gls{llm} (Claude) manually, \ie, by copy-pasting JGEA output within templated prompt and inserting \gls{llm} responses within JGEA experiment description file.
This way we are able to monitor in real time the correct execution of the interaction.

The execution of one interaction with a variant with feedback lasts on average \qty{20}{\minute}: $\approx$ \qty{10}{\percent} is taken by the \gls{llm} to respond (with no significant variance across feedback modalities), the remaining by the optimizer---we run it on a machine equipped with an Intel® Core™ i5 14500T vPro® CPU (\num{14} cores at \qty{4.8}{\giga\hertz}, with \num{20} threads) and \qty{16}{\giga\byte} of RAM.

The code for reproducing our experiments is available at \url{https:// redacted.for.review}.

\subsection{General effectiveness (\ref{item:rq1-general-effectiveness})}
\label{sec:experiments-general-effectiveness}
\begin{figure}
    \centering
    \tikzsetnextfilename{fitness-train-best}
    \begin{tikzpicture}[baseline=(current bounding box.center)]
        \readlist\colorlist{col11,col12,col13,col14,col15,col17}
        \begin{groupplot}[
            width=30mm,
            height=30mm,
            group style={
                group size=2 by 2,
                horizontal sep=8mm,
                vertical sep=2mm,
                xticklabels at=edge bottom,
                xlabels at=edge bottom,
                ylabels at=edge left
            },            
            scale only axis,            
            noinnerticks,
            scaled x ticks=base 10:-4,
            xtick scale label code/.code={}
        ]
            \nextgroupplot[title={Progression}, ylabel = {Training $f\left(s^\star;C_\text{train}\right)$}, gridded]
            \pgfplotstableread[]{data/train-train-line.txt}\data;
            \lineminmax[lcolor=col11]{\data}{}{step}{version0}{version0_min}{version0_max}
            \lineminmax[lcolor=col12]{\data}{}{step}{version1}{version1_min}{version1_max}
            \lineminmax[lcolor=col13]{\data}{}{step}{version2}{version2_min}{version2_max}
            \lineminmax[lcolor=col14]{\data}{}{step}{static_cl}{static_cl_min}{static_cl_max}
            \lineminmax[lcolor=col15]{\data}{}{step}{random_cl}{random_cl_min}{random_cl_max}
            \lineminmax[lcolor=col17]{\data}{}{step}{handmade}{handmade_min}{handmade_max}     
            
            \nextgroupplot[title={Final distribution}, ygridded, boxplot, boxplot/draw direction=y, ymax=0.62]
            \pgfplotstableread[]{data/train-train-box.txt}\data;
            \boxplot[bpcolor=col11]{\data}{version0};
            \boxplot[bpcolor=col12]{\data}{version1};
            \boxplot[bpcolor=col13]{\data}{version2};
            \boxplot[bpcolor=col14]{\data}{static_cl};
            \boxplot[bpcolor=col15]{\data}{random_cl};
            \boxplot[bpcolor=col17]{\data}{handmade};

            \boxplotcoloredstars{1}{0.55}{0,0,0,0,1,0}
            \boxplotcoloredstars{2}{0.57}{0,0,0,0,1,1}
            \boxplotcoloredstars{3}{0.55}{0,0,0,0,1,0}
            \boxplotcoloredstars{4}{0.57}{0,0,0,0,1,0}
            \boxplotcoloredstars{5}{0.55}{1,1,1,1,0,1}
            \boxplotcoloredstars{6}{0.57}{0,1,0,0,1,0}

            \nextgroupplot[ylabel={Fitness}, ylabel = {Gener.\ $f\left(s^\star;C_\text{test}\right)$}, gridded, xlabel={Stages}]
            \pgfplotstableread[]{data/train-test-line.txt}\data;
            \lineminmax[lcolor=col11]{\data}{}{step}{version0}{version0_min}{version0_max}
            \lineminmax[lcolor=col12]{\data}{}{step}{version1}{version1_min}{version1_max}
            \lineminmax[lcolor=col13]{\data}{}{step}{version2}{version2_min}{version2_max}
            \lineminmax[lcolor=col14]{\data}{}{step}{static_cl}{static_cl_min}{static_cl_max}
            \lineminmax[lcolor=col15]{\data}{}{step}{random_cl}{random_cl_min}{random_cl_max}
            \lineminmax[lcolor=col17]{\data}{}{step}{handmade}{handmade_min}{handmade_max}     
            
            \nextgroupplot[ygridded, boxplot, boxplot/draw direction=y, ymax=1]
            \pgfplotstableread[]{data/train-test-box.txt}\data;
            \boxplot[bpcolor=col11]{\data}{version0};
            \boxplot[bpcolor=col12]{\data}{version1};
            \boxplot[bpcolor=col13]{\data}{version2};
            \boxplot[bpcolor=col14]{\data}{static_cl};
            \boxplot[bpcolor=col15]{\data}{random_cl};
            \boxplot[bpcolor=col17]{\data}{handmade};

            \boxplotcoloredstars{1}{0.92}{0,1,1,0,1,0}
            \boxplotcoloredstars{2}{0.95}{1,0,0,1,1,0}
            \boxplotcoloredstars{3}{0.92}{1,0,0,1,1,0}
            \boxplotcoloredstars{4}{0.95}{0,1,1,0,1,1}
            \boxplotcoloredstars{5}{0.92}{1,1,1,1,0,1}
            \boxplotcoloredstars{6}{0.95}{0,0,0,1,1,0}
            
        \end{groupplot}
    \end{tikzpicture}
    \\
    \begin{tabular}{llll}
        W/ feedback: & 
        \addlegendimageintext{shaded legend image={col11}{solid}}
        \addlegendimageintext{boxplot legend image={col11}{solid}}
        \fN &
        \addlegendimageintext{shaded legend image={col12}{solid}}
        \addlegendimageintext{boxplot legend image={col12}{solid}}
        \fNP &
        \addlegendimageintext{shaded legend image={col13}{solid}}
        \addlegendimageintext{boxplot legend image={col13}{solid}}
        \fNPB \\
        W/o feedback: &
        \addlegendimageintext{shaded legend image={col14}{solid}}
        \addlegendimageintext{boxplot legend image={col14}{solid}}
        Static &
        \addlegendimageintext{shaded legend image={col15}{solid}}
        \addlegendimageintext{boxplot legend image={col15}{solid}}
        Random &
        \addlegendimageintext{shaded legend image={col17}{solid}}
        \addlegendimageintext{boxplot legend image={col17}{solid}}
        Expert \\
    \end{tabular}
    \vspace{-2mm}
    \caption{
        Progression (right plots) and final distribution (left plots) of the performance of the best policy $s^\star$ in the population measured on the train arenas (top plots) and on the test arenas (bottom plots), for the three interactive modalities and the three baselines.
        In the progression plots, the line corresponds to the median value across the \num{10} repetitions, the shaded area to the interquartile range.
        In the distribution plots, stars above the boxes show significant differences: \eg, an orange star over the red box indicates that \fN is significantly different than Random.
    }
    \label{fig:fitness-train-best}
\end{figure}
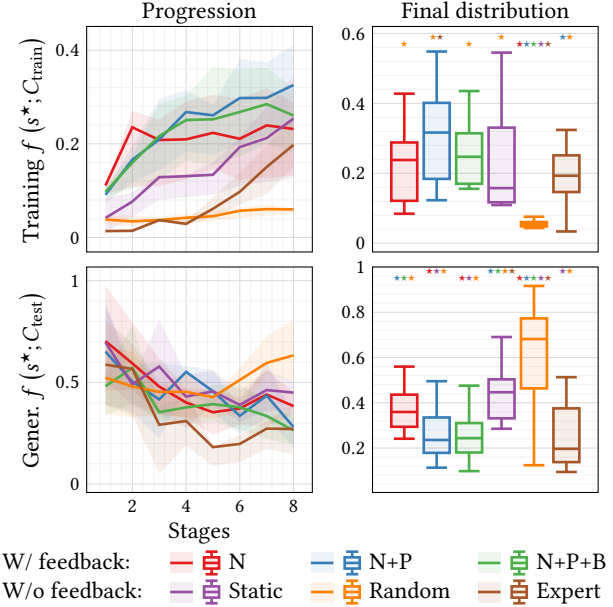

\Cref{fig:fitness-train-best} presents the learning dynamics across all methods over eight curriculum stages, showing both \emph{training performance} (top row) and test \emph{generalization} (bottom row).
Namely, the figure shows the value of $f$ (weighted average of final and average distance to target) computed for the best policy $s^\star$ in the population respectively on the train ($C_\text{train}$) or test ($C_\text{test}$) arenas---the lower, the better.
The left column shows progression across stages as line plots (median and interquartile range over ten repetitions), while the right column shows distribution of the final training and test performance.

The line plots reveal two patterns. 
For training performance, performance $f\left(s^\star;C_\text{train}\right)$ generally increases across stages for all methods except Random \addlegendimageintext{shaded legend image={col15}{solid}}, indicating that curricula progressively introduce harder cases as intended. 
For test performance, performance $f\left(s^\star;C_\text{test}\right)$ decreases across stages for all methods except Random \addlegendimageintext{shaded legend image={col15}{solid}}, demonstrating successful learning and generalization to unseen environments.

All \gls{llm}-generated curricula (active \fN, \fNP, \fNPB, and Static) significantly outperform the Random \addlegendimageintext{boxplot legend image={col15}{solid}} baseline and demonstrate clear learning. 
Random \addlegendimageintext{shaded legend image={col15}{solid}} does not show improvement in test performance across stages, suggesting that structured curriculum design (whether from the \gls{llm} or expert) is essential for effective learning. 
\gls{llm}-generated curricula \addlegendimageintext{shaded legend image={col11}{solid}} \addlegendimageintext{shaded legend image={col12}{solid}} \addlegendimageintext{shaded legend image={col13}{solid}} \addlegendimageintext{shaded legend image={col14}{solid}} successfully create meaningful learning progressions.

While all \gls{llm} methods show learning, none surpass the Expert \addlegendimageintext{boxplot legend image={col17}{solid}} curriculum in final performance. 
However, \fNP \addlegendimageintext{boxplot legend image={col12}{solid}} and \fNPB \addlegendimageintext{boxplot legend image={col13}{solid}} achieve performance statistically equivalent to the expert baseline (Mann-Whitney U test, $p>0.05$), indicating that \gls{llm}-assisted curriculum generation with appropriate feedback modalities can match expert-level design.

Two variants with feedback outperform Static generation.
\fNP \addlegendimageintext{boxplot legend image={col12}{solid}} and \fNPB \addlegendimageintext{boxplot legend image={col13}{solid}} achieve better test performance than Static \addlegendimageintext{boxplot legend image={col14}{solid}} (Mann-Whitney U test, all $p < 0.05$).
This demonstrates that providing iterative performance feedback enables the \gls{llm} to adapt curriculum difficulty and design more effective learning progressions for these versions.
These results validate the core hypothesis: active, feedback-informed curriculum generation produces better learning outcomes than static, one-shot curriculum design.

\paragraph{Feedback modality}
\fNP \addlegendimageintext{boxplot legend image={col12}{solid}} and \fNPB \addlegendimageintext{boxplot legend image={col13}{solid}} outperform \fN \addlegendimageintext{boxplot legend image={col11}{solid}}. 
Both progression-augmented (\fNP) and behavioral-augmented (\fNPB) feedback yield better test performance than numeric-only feedback (\fN) (Mann-Whitney U test, $p<0.05$).
The addition of the convergence plot visualization appears to provide the \gls{llm} with actionable insights about optimization dynamics, enabling more informed curriculum design (see, \eg, the last stage response of the \gls{llm} in the complete interaction shown in Appendix~\ref{sec:appendix-example}).

\fNP and \fNPB show no significant difference.
This suggests that while adding convergence visualization to numeric feedback improves performance, incorporating trajectory visualizations does not provide substantial additional benefits.
Analysis of the \gls{llm} \llm{understood} and \llm{reasoning} response fields reveals that it correctly interprets the trajectory visualizations and describes observed behaviors accurately. 
However, this additional information may also introduce prompt complexity without adding decision-relevant insights beyond what convergence plots already provide. 
The longer, more complex prompts in \fNPB may also introduce cognitive load that offsets potential benefits~\cite{nayab2024concise}.

\subsubsection{Generalization and \gls{qd}}
\label{sec:experiments-qd}
As we use a \gls{qd} optimizer (namely, \gls{me}), we obtain a set of diverse and good policies at the end of each optimization step, rather than just one good policy.
In particular, we use the final gap in $x$ and $y$ axes between the target and the robot position at the end of simulation.
Intuitively, different policies in the \gls{me} archive are hence diverse in the direction they approach the target from.
Indeed, we use these descriptors exactly because they can facilitate the finding of policies that truly generalize the target-reaching behavior beyond the examples provided by training arenas---\ie, to really solve the \gls{mt} 2D navigation task.

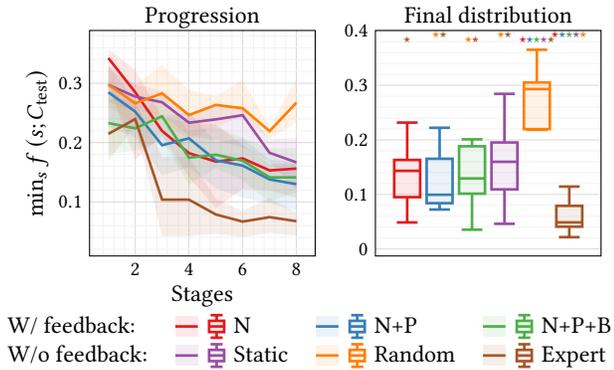
\begin{figure}
    \centering
    \tikzsetnextfilename{fitness-test-best}
    \begin{tikzpicture}[baseline=(current bounding box.center)]
        \readlist\colorlist{col11,col12,col13,col14,col15,col17}
        \begin{groupplot}[
            width=30mm,
            height=30mm,
            group style={
                group size=2 by 1,
                horizontal sep=8mm,
                vertical sep=2mm,
                xticklabels at=edge bottom,
                xlabels at=edge bottom,                
                ylabels at=edge left
            },
            scale only axis,            
            noinnerticks,
            scaled x ticks=base 10:-4,
            xtick scale label code/.code={},
            ylabel={$\min_s f\left(s;C_\text{test}\right)$}
        ]
            \nextgroupplot[title={Progression}, xlabel={Stages}, gridded]
            \pgfplotstableread[]{data/test-test-line.txt}\data;
            \lineminmax[lcolor=col11]{\data}{}{step}{version0}{version0_min}{version0_max}
            \lineminmax[lcolor=col12]{\data}{}{step}{version1}{version1_min}{version1_max}
            \lineminmax[lcolor=col13]{\data}{}{step}{version2}{version2_min}{version2_max}
            \lineminmax[lcolor=col14]{\data}{}{step}{static_cl}{static_cl_min}{static_cl_max}
            \lineminmax[lcolor=col15]{\data}{}{step}{random_cl}{random_cl_min}{random_cl_max}
            \lineminmax[lcolor=col17]{\data}{}{step}{handmade}{handmade_min}{handmade_max}    
            
            \nextgroupplot[title={Final distribution}, ygridded, boxplot, boxplot/draw direction=y]
            \pgfplotstableread[]{data/test-test-box.txt}\data;
            \boxplot[bpcolor=col11]{\data}{version0};
            \boxplot[bpcolor=col12]{\data}{version1};
            \boxplot[bpcolor=col13]{\data}{version2};
            \boxplot[bpcolor=col14]{\data}{static_cl};
            \boxplot[bpcolor=col15]{\data}{random_cl};
            \boxplot[bpcolor=col17]{\data}{handmade};

            \boxplotcoloredstars{1}{0.37}{0,0,0,0,0,1}
            \boxplotcoloredstars{2}{0.38}{0,0,0,0,1,1}
            \boxplotcoloredstars{3}{0.37}{0,0,0,0,1,1}
            \boxplotcoloredstars{4}{0.38}{0,0,0,0,1,1}
            \boxplotcoloredstars{5}{0.37}{1,1,1,1,0,1}
            \boxplotcoloredstars{6}{0.38}{1,1,1,1,1,0}            
            
        \end{groupplot}
    \end{tikzpicture} 
    \\
    \begin{tabular}{llll}
        W/ feedback: & 
        \addlegendimageintext{shaded legend image={col11}{solid}}
        \addlegendimageintext{boxplot legend image={col11}{solid}}
        \fN &
        \addlegendimageintext{shaded legend image={col12}{solid}}
        \addlegendimageintext{boxplot legend image={col12}{solid}}
        \fNP &
        \addlegendimageintext{shaded legend image={col13}{solid}}
        \addlegendimageintext{boxplot legend image={col13}{solid}}
        \fNPB \\
        W/o feedback: &
        \addlegendimageintext{shaded legend image={col14}{solid}}
        \addlegendimageintext{boxplot legend image={col14}{solid}}
        Static &
        \addlegendimageintext{shaded legend image={col15}{solid}}
        \addlegendimageintext{boxplot legend image={col15}{solid}}
        Random &
        \addlegendimageintext{shaded legend image={col17}{solid}}
        \addlegendimageintext{boxplot legend image={col17}{solid}}
        Expert \\
    \end{tabular}
    \vspace{-2mm}
    \caption{
        Progression (right plots) and final distribution (left plots) of the performance of the policy $s^\star$ with the best performance on the test arenas, for the three interactive modalities and the three baselines.        
    }
    \label{fig:fitness-test-best}
\end{figure}

To demonstrate that \gls{me} actually exploits this potential in achieving generalization, we show in \Cref{fig:fitness-test-best} the performance in the test arenas of the best solution in the \gls{me} archive, \ie, the progression and distribution of $\min_s f\left(s;C_\text{test}\right)$.

It can be seen that for all methods there is at least one policy in the archive which scores much better than the best solution chosen based on the performance on the training arenas: \eg, the median $\min_s f\left(s;C_\text{test}\right)$ of the final policy obtained with the \fNPB feedback is $\approx 0.13$ (\addlegendimageintext{boxplot legend image={col13}{solid}} in \Cref{fig:fitness-test-best}), while the median of $f\left(s^\star;C_\text{test}\right)$ is $\approx 0.22$ (\addlegendimageintext{boxplot legend image={col13}{solid}} in \Cref{fig:fitness-train-best}).
However, the largest difference is in the populations evolved with the Expert curriculum, particularly visible by comparing the progression of the index (\addlegendimageintext{shaded legend image={col17}{solid}}).

\subsection{Progressive \vs batch cases (\ref{item:rq2-progressiveness})}
\label{sec:experiments-progressive-vs-static}
To verify that the \gls{llm}-generated cases form a meaningful curriculum progression rather than arbitrary arena collections, we compare training with progressive administration of cases against training with \emph{batch} administration of all cases.
For each of our variant (\fN, \fNP, \fNPB), we take each of the $10$ curricula obtaned with feedback and perform a single optimization round with a $C_\text{train}$ containing all the eight curriculum cases.
To ensure fair comparison in terms of total episodes, we run the batch training longer, with \num{40000} fitness evaluations as stopping criterion: this way we perform the same number of episodes in both conditions ($8 \cdot \num{40000}=\sum_{i=1}^{i=8} \num{10000}$).

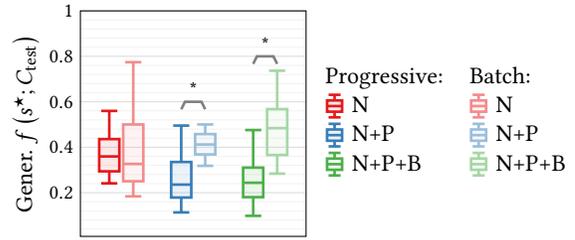
\begin{figure}
    \centering
    \tikzsetnextfilename{progressive-vs-batch}
    \begin{tikzpicture}[baseline=(current bounding box.center)]
        \begin{axis}[
            width=30mm,
            height=30mm,
            scale only axis,            
            noinnerticks,
            ygridded,
            boxplot,
            boxplot/draw direction=y,
            groupedboxplot={2}{0.28},
            ylabel=Gener.\ $f\left(s^\star;C_\text{test}\right)$,
            ymax=1
        ]
            \pgfplotstableread[]{data/cl-train-test-box.txt}\data;
            \boxplot[bpcolor=col11]{\data}{version0};
            \boxplot[bpcolor=col11!50]{\data}{version0_sallstages};
            \boxplot[bpcolor=col12]{\data}{version1};
            \boxplot[bpcolor=col12!50]{\data}{version1_sallstages};
            \boxplot[bpcolor=col13]{\data}{version2};
            \boxplot[bpcolor=col13!50]{\data}{version2_sallstages};

            \groupedpvalue{2}{2}{1}{2}{2}{0.6}{*}
            \groupedpvalue{2}{3}{1}{3}{2}{0.8}{*}
            
        \end{axis}
    \end{tikzpicture}
    \vspace{-5mm}
    \begin{tabular}{ll}
        Progressive: & Batch: \\ 
        \addlegendimageintext{boxplot legend image={col11}{solid}} \fN &
        \addlegendimageintext{boxplot legend image={col11!50}{solid}} \fN \\
        \addlegendimageintext{boxplot legend image={col12}{solid}} \fNP &
        \addlegendimageintext{boxplot legend image={col12!50}{solid}} \fNP \\
        \addlegendimageintext{boxplot legend image={col13}{solid}} \fNPB &
        \addlegendimageintext{boxplot legend image={col13!50}{solid}} \fNPB \\
    \end{tabular}
    \caption{
        Final distribution of the performance of the best policy $s^\star$ in the population measured on the test arenas for each modality with progressive and batch administration.
        An arch over a pair of boxes denotes statistically significant difference.
    }
    \label{fig:progressive-vs-batch}
\end{figure}

We show the performance of the obtained policies in \Cref{fig:progressive-vs-batch}.
The results demonstrate that \fNP and \fNPB truly benefit from curriculum structure: progressive training \addlegendimageintext{boxplot legend image={col12}{solid}} \addlegendimageintext{boxplot legend image={col13}{solid}} significantly outperforms the batch variants \addlegendimageintext{boxplot legend image={col12!50}{solid}} \addlegendimageintext{boxplot legend image={col13!50}{solid}} (Mann-Whitney U test, $p < 0.05$).
Conversely, there is no difference from progressive \addlegendimageintext{boxplot legend image={col11}{solid}} and batch \addlegendimageintext{boxplot legend image={col11!50}{solid}} administration of curricula obtained with \fN.
This confirms that the augmented feedback provided by progression plots and behavior visualization allows the \gls{llm} to generate sequences being genuine curricula, \ie, such that case ordering and gradual accumulation facilitate learning, rather than simply diverse training sets.

\subsection{Analyses of \gls{llm} responses (\ref{item:rq3-reasoning})}
\label{sec:llm-reasoning-insights}
\begin{figure}
    \centering
    \pgfsetlayers{background,main}
    \begin{tikzpicture}[
        prompt/.style={
            font=\linespread{0.8}\ttfamily\scriptsize,
            text=darkgray,            
            align=left,
            anchor=north
        },
        msg/.style={
            rounded corners=1mm,
            fill=yellow!10,
            draw=black            
        }
    ]
        \node[label={[yshift=-2mm]above:{Arenas (and $s^\star$ behavior) after stage 3 \numcirc{1} optimization}}] (arenas3) at (0,0) {
            \includegraphics[width=4.5cm]{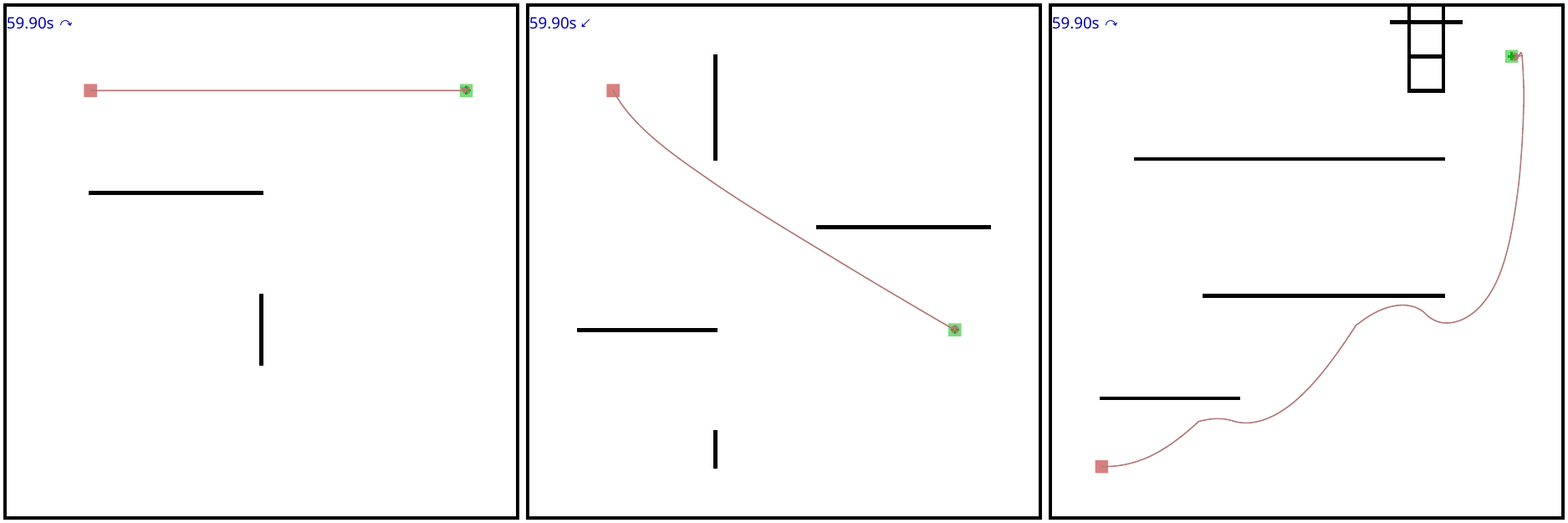}
        };
        \node[prompt,text width=7.5cm] (prompt-metric) at ($(arenas3.south)+(0,-0.7)$) {
            \textbf{Metrics}\\
            - quality of the best policy on the last stage : {0.023}\\
            - quality of the best policy on previous stages : {0.003, 0.003, 0.023}\\
            - quality on average of all stages: {0.0097}
        };
        \node[prompt,anchor=north west] (prompt-plot) at (prompt-metric.south west) {
            \textbf{Progression of the quality}\\
            \includegraphics[width=3cm]{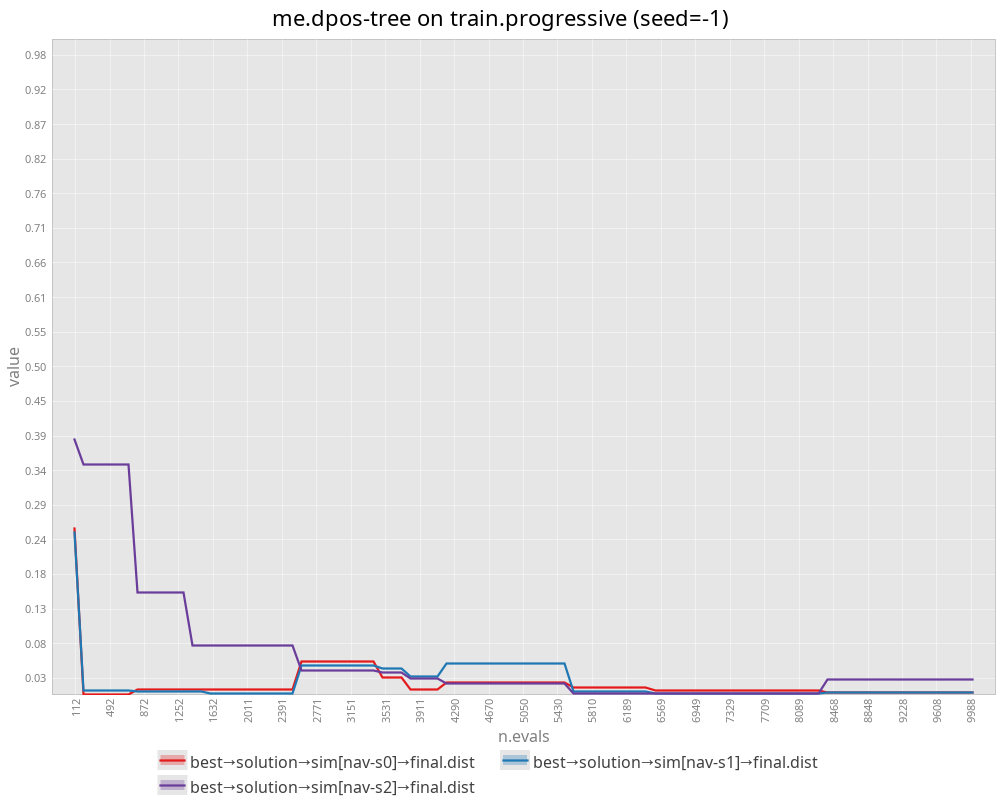}
        };
        \node[prompt,anchor=north west] (prompt-behavior) at (prompt-plot.north east) {
            \textbf{Policy behavior}\\
            \includegraphics[width=3cm]{data/good-run/imgs/s2-train.best-train.trajs_svg-raw.pdf}
        };
        \begin{scope}[on background layer]
            \node[msg,label={[yshift=-1mm]above:{\numcirc{2} Feedback after stage 3}},fit=(prompt-metric) (prompt-plot) (prompt-behavior)] (feedback3) {};
        \end{scope}

        \node[msg,prompt,text width=7.5cm,label={[yshift=-1mm]above:{\numcirc{3} Response 3}}] (response3) at ($(feedback3.south)+(0,-0.7)$) {
            \textbf{understood}\\
            Stage 3 achieved quality 0.023 after starting at 0.37 and converging by evaluation 2000; the trajectory shows a smooth S-curve path from left-middle start navigating upward through horizontal wall barriers to reach the upper-right target, demonstrating successful zigzag maneuvering.\\
            \textbf{reasoning}\\
            With consistently strong performance across stages, this next arena introduces a vertical corridor structure on the left side with the target inside it, plus additional horizontal barriers in the lower-right area where the start is located, requiring the policy to navigate through a more constrained passage while maintaining path diversity from previous horizontal-dominated layouts.
        };

        \node[anchor=north,label={[yshift=-2mm]above:{Arenas (and $s^\star$ behavior) after stage 4 \numcirc{1} optimization}}] (arenas4) at ($(response3.south)+(0,-0.7)$) {
            \includegraphics[width=6cm]{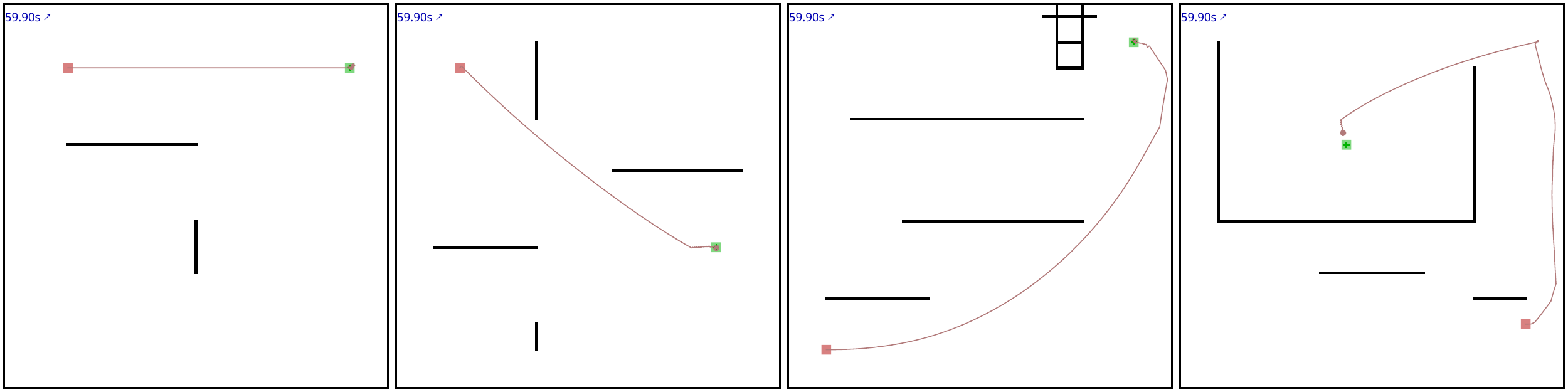}
        };
    \end{tikzpicture}    
    \caption{
        A visualization of a chunk of one run of the interaction with the \fNPB feedback: steps \protect\numcirc{1}\protect\numcirc{2}\protect\numcirc{3} of stage 3 and step \protect\numcirc{1} of stage 4.
        The complete interaction including these stages is shown in Appendix~\ref{sec:appendix-example}.
    }
    \label{fig:one-run}
\end{figure}

\Cref{fig:one-run} illustrates one stage of one interaction loop using the \fNPB modality, where $3$ cases have already been generated and evaluated.
The \gls{llm} receives numeric metrics, convergence plots, and trajectory visualizations, then produces the next case along with explanatory text.
The complete sequence of prompts and \gls{llm} responses across all stages of this interaction is provided in Appendix~\ref{sec:appendix-example}.

By observing the \llm{understood} field of the \gls{llm} response in \Cref{fig:one-run}, it can be seen that the \gls{llm} demonstrates basic competence in interpreting the provided visualizations. 
It correctly identifies convergence patterns from the progression plots, including the starting quality value, approximate evaluation count at convergence, and relative convergence speed. 
For trajectory plots, it accurately locates start and target positions and describes the general path shape (\eg, \llm{smooth curve path}, \llm{diagonal path}). 
The \gls{llm} also correctly uses the legend to associate performance curves with specific cases in multi-line plots.

From the \llm{reasoning} field in \Cref{fig:one-run}, we see that the \gls{llm} is able to exploit the understanding of the feedback for providing and motivating the new case. 
By looking more broadly at other interactions, we see that when policies achieve low quality scores consistently, the \gls{llm} introduces additional obstacles, increases spatial complexity, or varies object positions to maintain challenge progression. 
Conversely, when performance degrades significantly, it explicitly acknowledges the difficulty spike and generates simpler arenas.
Comparing the \gls{llm} textual descriptions of intended arena layouts with the actual 2D environments rendered from its character strings reveals reasonable consistency; the generated structures generally match the stated design intentions.

These observations suggest that the \gls{llm} can extract relevant information from multimodal feedback and adjust curriculum difficulty accordingly, though its designs represent functional rather than optimal curriculum strategies.

\glsresetall

\section{Concluding remarks}
\label{sec:conclusions}
We introduced an interactive \gls{llm}-assisted framework that automates curriculum generation for \gls{mt} evolutionary policy search by leveraging real-time optimization feedback.
Experiments on 2D robotic navigation demonstrated that interactive \gls{llm} curricula outperform static \gls{llm} and random baselines, with progression-augmented feedback (\fNP) and behavior-augmented feedback (\fNPB) achieving performance comparable to expert-designed curricula.

Our findings reveal that feedback richness matters: adding convergence plots to numeric metrics improves curriculum quality, though behavioral visualizations offer more modest incremental gains.
Analysis of \gls{llm} responses confirmed meaningful interpretation of multimodal feedback and adaptive difficulty calibration.
Crucially, curricula generated with richer feedback function as genuine learning progressions, as sequential administration outperforms batch training.

This framework is agnostic to optimizer, policy representation, and problem domain, opening pathways for broader evolutionary robotics applications.
Future work should investigate scaling to complex robotic tasks, alternative \glspl{llm}, and whether incorporating the evolved policies or generalization metrics (\eg, validation set performance) into the feedback loop enables \glspl{llm} to better anticipate curriculum impact on unseen cases.
This work demonstrates that \glspl{llm} can effectively automate curriculum design for embodied \gls{ai}, lowering the expertise barrier for \gls{mt} learning in evolutionary systems.

%acks: cost action

\bibliographystyle{ACM-Reference-Format}
\bibliography{bibliography}

\newpage
\appendix
\section{Example of a complete interaction}
\label{sec:appendix-example}
We show here a complete execution of our pipeline with the \fNPB feedback.
Namely, we show the \numcirc{0} contextualization prompt, which the optimizer sends before the first stage, and the three key steps for every stage: \numcirc{1} the optimization, \numcirc{2} the feedback sent by the optimizer to the \gls{llm}, and \numcirc{3} the \gls{llm} response.

For the \numcirc{1} optimization, we show the behavior (\ie, the trajectory) of the best policy at the end of the evolution on the train cases (\ie, arenas) of the current stage and the plot of the progression during the optimization of the performance of the best policy on the train cases.
We also show the behavior of the same individual on the test cases, for showing how the generalization of the policy improves during the entire process: we recall, however, that this behavior and, more broadly, the performance on the test cases are never known by the optimizer and the \gls{llm}.

For the \numcirc{2} feedback, we show the exact prompt the optimizer submits to the \gls{llm}: in the actual interaction, we also attached the image of the behavior of the best policy on the train cases and the plot of the progression of the performance (the very same images shown here in the preceding optimization step).

For the \numcirc{3} response, we show the exact response given by the \gls{llm}.

\subsection{Before stage 1}
\subsubsection{\numcirc{0} Contextualization}\hfill
\begin{lstlisting}[inputencoding=utf8]
CONTRACT: STAGE-GRID v1.0
You are a STAGE DESIGN assistant for CURRICULUM LEARNING in evolutionary control. I am optimizing the policy of a differential drive robot in a 2D arena. The robot must START at 's', REACH the target 't', and CANNOT PASS over walls 'w'. I am expecting to obtain via curriculum learning, a general policy for ANY arena, I will evaluate GENERALIZATION with test arenas. 
## PURPOSE & USE
You will provide STAGES (arenas) for a sequential curriculum learning process. Each stage will be used to assess candidate policies during optimization, which proceeds in sequential STAGES (curriculum steps). You DO NOT assist with optimization; you only output the NEXT STAGE and a brief explanation (“understood” and “reasoning”).
## OBJECTIVE QUALITY
I assess candidate policies on the stages you provide via an objective score called "quality". The goal is to minimize the quality score (distance to target). While achieving quality < 0.1 represents optimal performance, policies with scores below 0.3 also demonstrate solid capabilities and meaningful progress.
## PROVIDED FEEDBACK
Before asking for each new stage, I will provide some feedback about the previous stage and the progress in general. Namely, I’ll provide:
- quality of the best policy on the last stage: {{current_quality}}        // lower = better
- quality of the best policy on previous stages : {{history_summary}} // compact stats over recent stages/qualities
- quality on average of all stages: {{average_quality}}        // lower = better
- a PLOT (for your reference, not to be produced) showing the evolution of the best individual's quality over the number of evaluations for each stage (s0, s1, s2, …). It indicates how quickly and how well the population improved during training. Use this trend to assess the current stage and adjust the next stage’s complexity accordingly.
- a second plot (for your reference, not to be produced) showing the ARENA (previous stage) and the TRAJECTORY of the best individual. It includes the start (pink square), target (green square), walls (black segments), and the trajectory (pink line). Use what you observe in this image to analyze the arena’s layout, difficulty, and robot behavior before generating the next stage. Treat it as a visual reference for the previous arena’s layout and the robot’s trajectory, this is very important information. This plot should help you understand the MAPPING between the STRING named grid and the VISUAL LAYOUT of the arena, USE this understanding. Use it to infer spatial relationships, but you don’t need to describe it verbatim, simply integrate your observations naturally into your explanation in “understood”. 
- When analyzing the trajectory plot, note which wall configurations created meaningful challenges versus simple obstacles. Use successful patterns to inform varied new designs, not repetitive ones.
## WHAT TO RETURN
Return ONE JSON object with the following keys:
- "grid": the next stage as a string encoding a matrix. The grid must follow the GRID FORMAT rules exactly (15×15, rows joined by |, cells in {w,s,t,e}).
- "understood": a string with your one-sentence interpretation of all the feedback I provided. (generic on first call). INCLUDING YOUR INTERPRETATION OF BOTH PLOTS (briefly summarize key visual observations from the image, such as the robot’s path shape, proximity to walls, or how it approached the target)
- "reasoning": a string with a short motivation for why this stage is appropriate as the next step in a curriculum.
## GRID FORMAT
- The grid is a single string with vertical bar ('|') as separator between rows.
- Size: 15 rows and 15 columns. All the rows have to have the SAME LENGTH (15).
- Allowed characters:
  - w = wall
  - s = starting position (exactly once)
  - t = target position (exactly once)
  - e = empty cell
Do not use '|' as a cell character. Allowed cells are only {w, s, t, e}.
## HARD CONSTRAINTS (must be logically satisfied before output)
  - Exactly one 's' and one 't'. 
  - There must be AT LEAST ONE CONTINUOUS OPEN PATH of  'e' cells connecting  's' to 't'!!! (walls 'w' block movement). INTERNAL VALIDATION PROCEDURE: Simulate a 4-connected flood-fill or BFS (moves: up/down/left/right only) starting from 's'. Confirm that 't' can be reached without crossing any 'w'. Only grids that pass this internal check are valid.
  - VARY the positions of 's' and 't'  across different stages.
  - All the rows HAVE TO have the SAME LENGTH.
  - Don’t include frames in the grid, it is provided automatically by default. 
Output only after rechecking that all HARD CONSTRAINTS are met.
## VALIDATION CHECK (must pass before output)
Before outputting, verify the given hard constraints were met.
If any condition fails, regenerate the grid and recheck.
Only output once all checks pass.
## OUTPUT FORMAT
CRITICAL: Each row = EXACTLY 15 characters. COUNT before output.
{
  "grid": "<grid-string-with-rows-joined-by-|>",
  "understood": "<your brief interpretation (generic on first call)>",
  "reasoning": "<why this stage is a sensible next step>"
}
## COMPLETE EXAMPLE (DO NOT COPY LITERALLY; FOLLOW THE STRUCTURE)
{
  "grid": "eeeeeeeeeeeeeee|eeeeeeeeeeeeeee|eeeeeeeeeeeeeee|eeeeeeeeteeeeee|eeeeeeeeeeeeeee|wwwwwwwwweeeeee|eeeeeeeeeeeeeee|eeeeeeeeeeeeeee|eeeeeeeeeeeeeee|eeeeeeewwwwwwww|eeeeeeeeeeeeeee|eseeeeeeeeeeeee|eeeeeeeeeeeeeee|wwwwweeeeeeeeee|eeeeeeeeeeeeeee",
  "understood": "No prior feedback provided; initializing curriculum with a solvable maze.",
  "reasoning": "Start and target are well-separated with multiple corridors; it supports early exploration while ensuring reachability."
}
## FINAL INSTRUCTION
Return ONLY a valid JSON object as specified above. No extra text, no code fences.
\end{lstlisting}
\subsubsection{\numcirc{3} Response 0}\hfill
\begin{lstlisting}[inputencoding=utf8]
{
  "grid": "eeeeeeeeeeeeeee|eeeeeeeeeeeeeee|eeseeeeeeeeeete|eeeeeeeeeeeeeee|eeeeeeeeeeeeeee|eewwwwwweeeeeee|eeeeeeeeeeeeeee|eeeeeeeeeeeeeee|eeeeeeeweeeeeee|eeeeeeeweeeeeee|eeeeeeeweeeeeee|eeeeeeeeeeeeeee|eeeeeeeeeeeeeee|eeeeeeeeeeeeeee|eeeeeeeeeeeeeee",
  "understood": "No prior feedback provided; initializing the curriculum with a simple yet structured arena to establish baseline performance.",
  "reasoning": "This first stage places start and target on the same row with moderate separation, introducing a single L-shaped wall obstacle that requires basic navigation around it. The layout is deliberately simple to allow the initial population to learn fundamental movement and target-seeking behaviors while ensuring a clear solvable path exists."
}
\end{lstlisting}

%\foreach \i in {0,1,2,3,4,5,6}{
%    \newpage
%    \pgfmathtruncatemacro{\trainwidth}{1+\i}
%    \subsection{Stage \trainwidth}
%    
%    \subsubsection{\numcirc{1} Optimization}\hfill\\
%    \includegraphics[width=\trainwidth cm]{data/good-run/imgs/s\i-train.best-train.trajs_svg-raw.pdf}\\
%    \includegraphics[width=6cm]{data/good-run/imgs/s\i-train.best-test.trajs_svg-raw.pdf}\\
%    \includegraphics[width=\linewidth]{data/good-run/imgs/s\i-PROGRESSION-train.best-train.ds.png}
    
%    \subsubsection{\numcirc{2} Feedback}\hfill
%    \lstinputlisting[inputencoding=utf8]{data/good-run/msgs/feedback\i.txt}
    
%    \subsubsection{\numcirc{3} Response}\hfill
%    \lstinputlisting[inputencoding=utf8]{data/good-run/msgs/response\i.txt}
%}

\newpage
\subsection{Stage 1}

\subsubsection{\numcirc{1} Optimization}\hfill\\
\includegraphics[width=1cm]{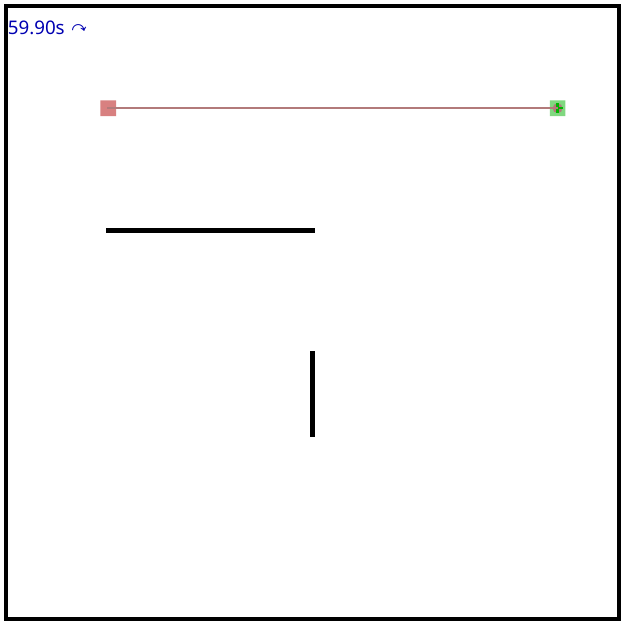}\\
\includegraphics[width=6cm]{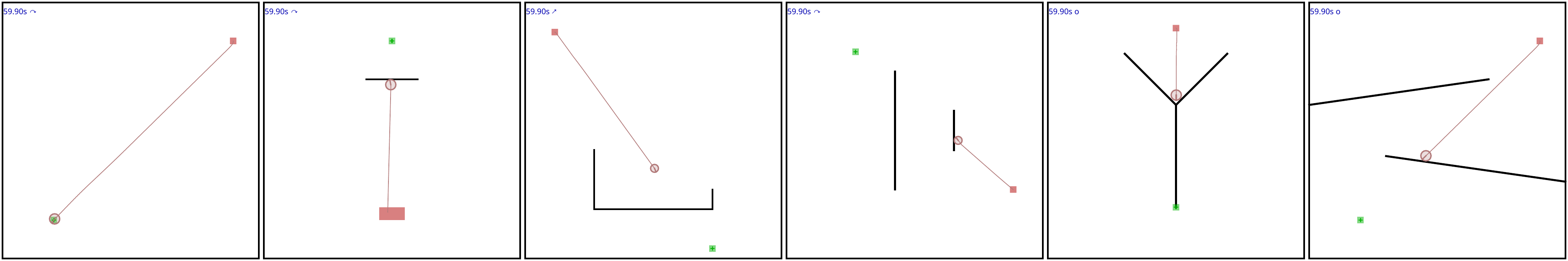}\\
\includegraphics[width=\linewidth]{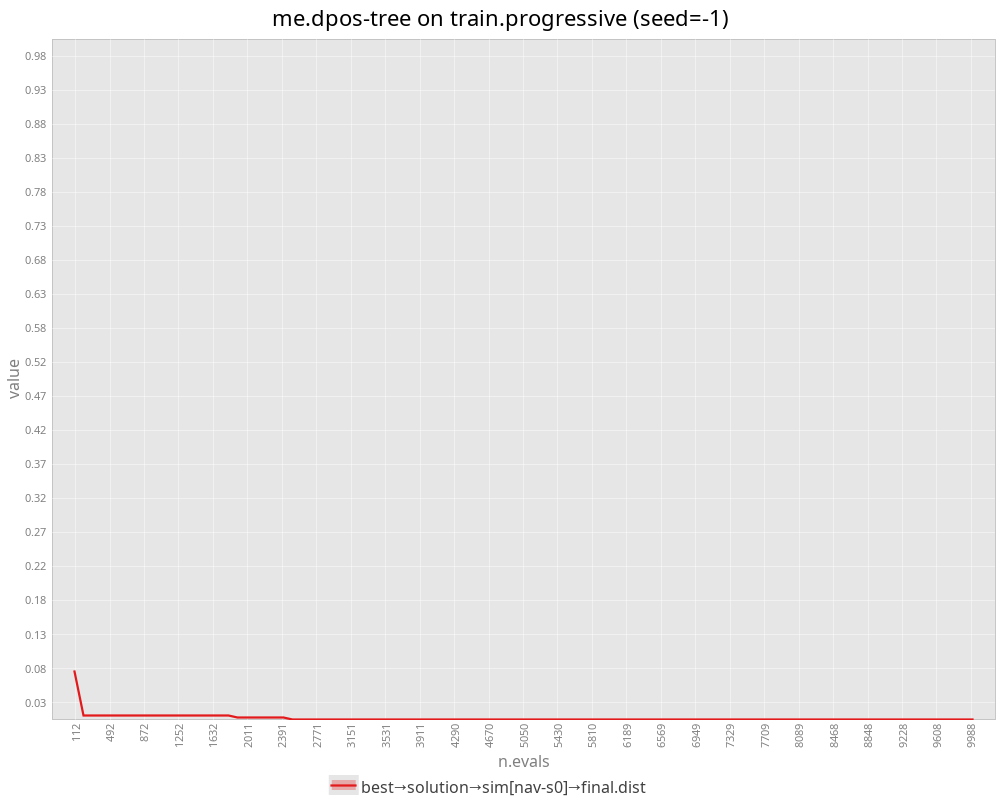}

\subsubsection{\numcirc{2} Feedback}\hfill
\begin{lstlisting}[inputencoding=utf8]
CONTRACT: STAGE-GRID v1.0 applies.
## PROVIDED FEEDBACK
- quality of the best policy on the last stage : {0.000}        
- quality of the best policy on previous stages : {0.000}
- quality on average of all stages: {0.000}
## REMINDER (do not change)
- Grid: 15x15; rows joined by '|' (do not use '|' as a cell).
- Size: exactly 15 rows × 15 columns; every row must be length 15. 
- Alphabet: {w, s, t, e}; exactly one 's' and one 't'.
- Variety: vary the positions of 's' and 't' across stages.
- Reachability: at least one 4-connected path of 'e' from 's' to 't' (no diagonals).
- Include: "understood" and "reasoning".
Don’t forget the constraints and validation check. If any output would violate CONTRACT: STAGE-GRID v1.0, regenerate internally and only emit a valid JSON object.
- the two plots are attached. The two plots are for reference only; do not reproduce them in the output.
## ROW LENGTH VALIDATION (CRITICAL)
1. Each row = exactly 15 characters
2. Total of 15 rows
3. Validation method: COUNT each row before output
4. If count ≠ 15 → REJECT and regenerate entire grid
\end{lstlisting}

\subsubsection{\numcirc{3} Response}\hfill
\begin{lstlisting}[inputencoding=utf8]
{
  "grid": "eeeeeeeeeeeeeee|eeeeeweeeeeeeee|eeseeweeeeeeeee|eeeeeweeeeeeeee|eeeeeweeeeeeeee|eeeeeeeeeeeeeee|eeeeeeeewwwwwwe|eeeeeeeeeeeeeee|eeeeeeeeeeeeeee|ewwwwweeeeeetee|eeeeeeeeeeeeeee|eeeeeeeeeeeeeee|eeeeeweeeeeeeee|eeeeeweeeeeeeee|eeeeeeeeeeeeeee",
  "understood": "Stage 0 achieved optimal quality (0.000) rapidly within 200 evaluations; the trajectory plot shows a nearly direct path from left-side start to right-side target with minimal deviation around the simple L-shaped walls.",
  "reasoning": "Since the previous stage was solved perfectly and quickly, this stage introduces three distinct wall obstacles forming partial barriers at different vertical positions, requiring the robot to navigate around multiple corners while maintaining a clear solvable path from the left-middle start to the right-middle target."
}
\end{lstlisting}

\newpage
\subsection{Stage 2}

\subsubsection{\numcirc{1} Optimization}\hfill\\
\includegraphics[width=2cm]{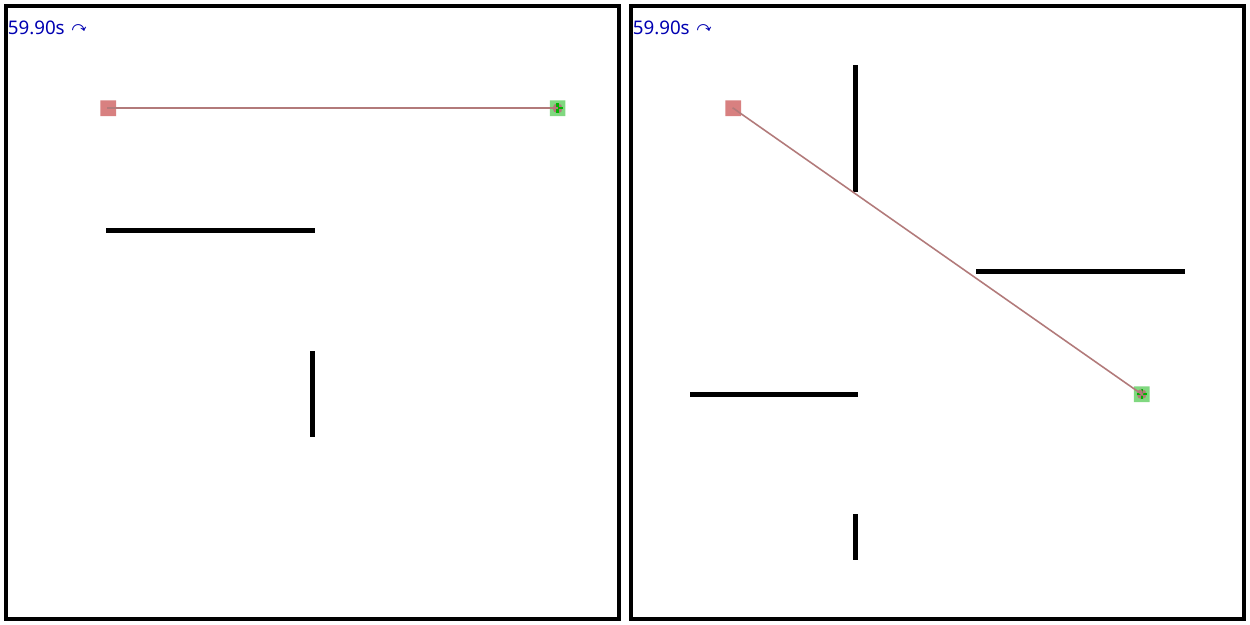}\\
\includegraphics[width=6cm]{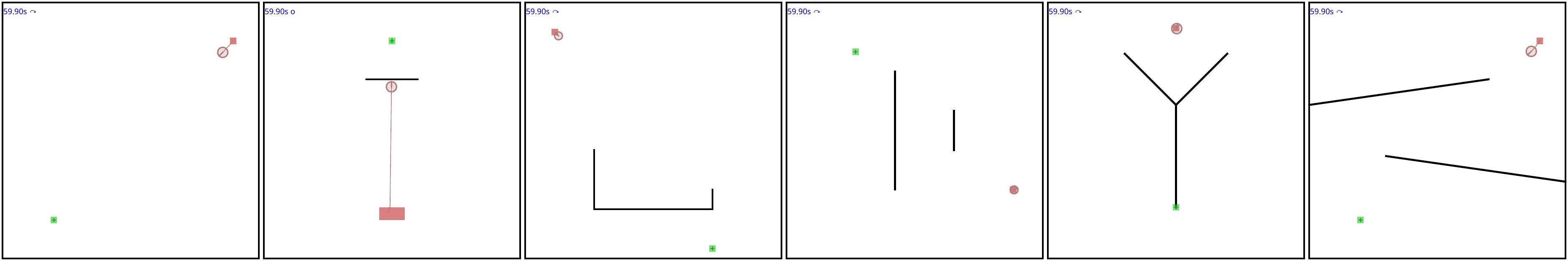}\\
\includegraphics[width=\linewidth]{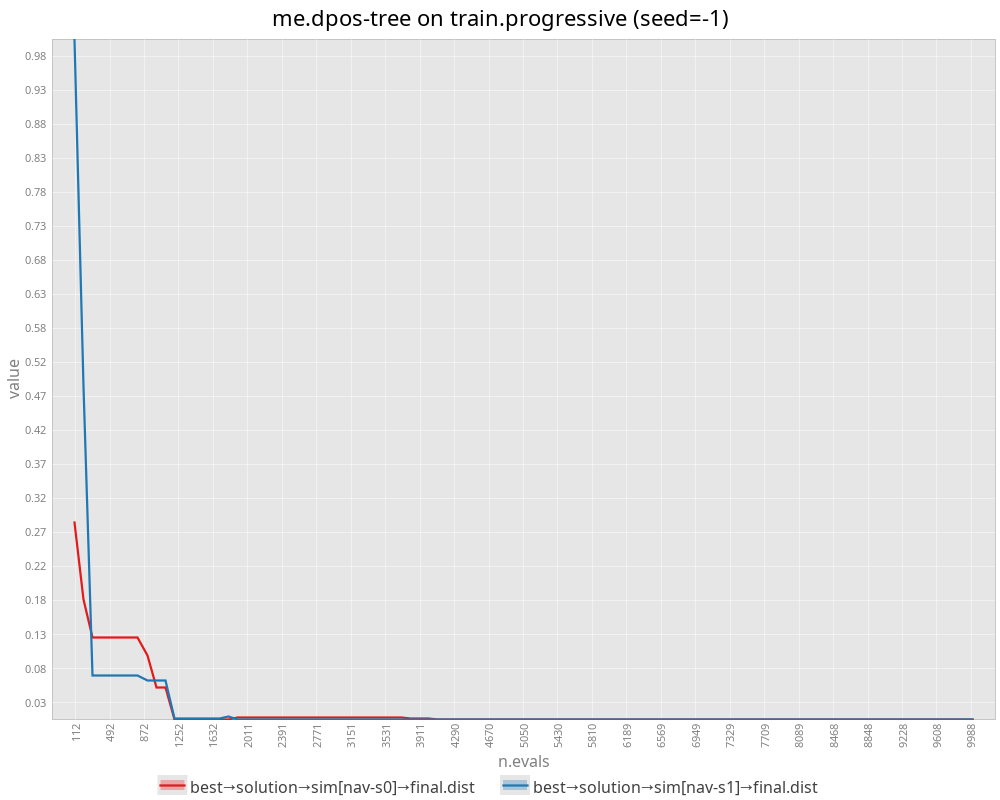}

\subsubsection{\numcirc{2} Feedback}\hfill
\begin{lstlisting}[inputencoding=utf8]
CONTRACT: STAGE-GRID v1.0 applies.
## PROVIDED FEEDBACK
- quality of the best policy on the last stage : {0.000}        
- quality of the best policy on previous stages : {0.000, 0.000}
- quality on average of all stages: {0.000}
## REMINDER (do not change)
- Grid: 15x15; rows joined by '|' (do not use '|' as a cell).
- Size: exactly 15 rows × 15 columns; every row must be length 15. 
- Alphabet: {w, s, t, e}; exactly one 's' and one 't'.
- Variety: vary the positions of 's' and 't' across stages.
- Reachability: at least one 4-connected path of 'e' from 's' to 't' (no diagonals).
- Include: "understood" and "reasoning".
Don’t forget the constraints and validation check. If any output would violate CONTRACT: STAGE-GRID v1.0, regenerate internally and only emit a valid JSON object.
- the two plots are attached. The two plots are for reference only; do not reproduce them in the output.
## ROW LENGTH VALIDATION (CRITICAL)
1. Each row = exactly 15 characters
2. Total of 15 rows
3. Validation method: COUNT each row before output
4. If count ≠ 15 → REJECT and regenerate entire grid
\end{lstlisting}

\subsubsection{\numcirc{3} Response}\hfill
\begin{lstlisting}[inputencoding=utf8]
{
  "grid": "eeeeeeeeeewweee|eeeeeeeeeewwete|eeeeeeeeeewweee|eeeeeeeeeeeeeee|eewwwwwwwwwweee|eeeeeeeeeeeeeee|eeeeeeeeeeeeeee|eeeeeeeeeeeeeee|eeeewwwwwwwweee|eeeeeeeeeeeeeee|eeeeeeeeeeeeeee|ewwwwweeeeeeeee|eeeeeeeeeeeeeee|eseeeeeeeeeeeee|eeeeeeeeeeeeeee",
  "understood": "Both stages achieved optimal quality (0.000); stage 1's progression shows initial struggle (starting near 1.0) before converging around evaluation 800, and the trajectory displays a smooth diagonal path from upper-left to lower-right navigating around the three scattered wall segments.",
  "reasoning": "With two consecutive optimal solutions, this stage introduces a more complex multi-barrier layout with three horizontal wall segments at different vertical levels (top, middle, lower-middle) creating a zigzag navigation requirement from bottom-left start to top-right target, testing the policy's ability to handle sequential obstacle avoidance."
}
\end{lstlisting}

\newpage
\subsection{Stage 3}

\subsubsection{\numcirc{1} Optimization}\hfill\\
\includegraphics[width=3cm]{data/good-run/imgs/s2-train.best-train.trajs_svg-raw.pdf}\\
\includegraphics[width=6cm]{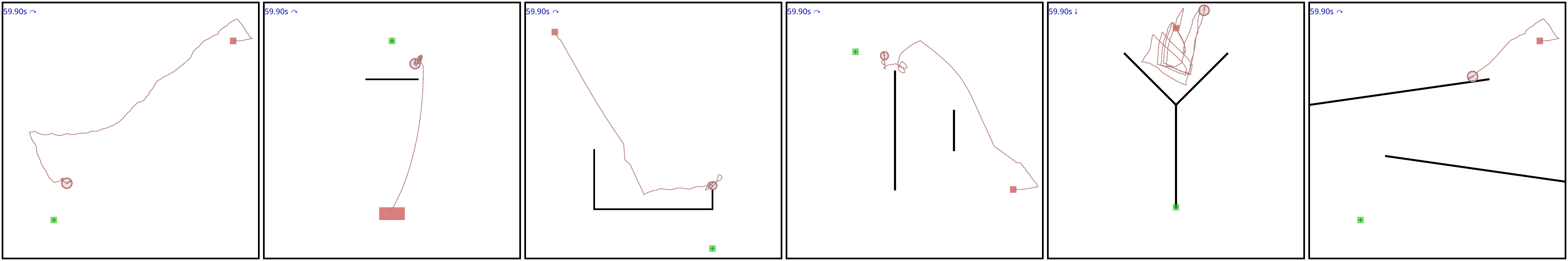}\\
\includegraphics[width=\linewidth]{data/good-run/imgs/s2-PROGRESSION-train.best-train.ds.png}

\subsubsection{\numcirc{2} Feedback}\hfill
\begin{lstlisting}[inputencoding=utf8]
CONTRACT: STAGE-GRID v1.0 applies.
## PROVIDED FEEDBACK
- quality of the best policy on the last stage : {0.023}        
- quality of the best policy on previous stages : {0.003, 0.003, 0.023}
- quality on average of all stages: {0.0097}
## REMINDER (do not change)
- Grid: 15x15; rows joined by '|' (do not use '|' as a cell).
- Size: exactly 15 rows × 15 columns; every row must be length 15. 
- Alphabet: {w, s, t, e}; exactly one 's' and one 't'.
- Variety: vary the positions of 's' and 't' across stages.
- Reachability: at least one 4-connected path of 'e' from 's' to 't' (no diagonals).
- Include: "understood" and "reasoning".
Don’t forget the constraints and validation check. If any output would violate CONTRACT: STAGE-GRID v1.0, regenerate internally and only emit a valid JSON object.
- the two plots are attached. The two plots are for reference only; do not reproduce them in the output.
## ROW LENGTH VALIDATION (CRITICAL)
1. Each row = exactly 15 characters
2. Total of 15 rows
3. Validation method: COUNT each row before output
4. If count ≠ 15 → REJECT and regenerate entire grid
\end{lstlisting}

\subsubsection{\numcirc{3} Response}\hfill
\begin{lstlisting}[inputencoding=utf8]
{
  "grid": "eeeeeeeeeeeeeee|eweeeeeeeeeeeee|eweeeeeeeeeweee|eweeeeeeeeeweee|eweeeeeeeeeweee|eweeeeteeeeweee|eweeeeeeeeeweee|eweeeeeeeeeweee|ewwwwwwwwwwweee|eeeeeeeeeeeeeee|eeeeewwwwweeeee|eeeeeeeeeeewwwe|eeeeeeeeeeeeese|eeeeeeeeeeeeeee|eeeeeeeeeeeeeee",
  "understood": "Stage 2 achieved quality 0.023 after starting at 0.37 and converging by evaluation 2000; the trajectory shows a smooth S-curve path from left-middle start navigating upward through horizontal wall barriers to reach the upper-right target, demonstrating successful zigzag maneuvering.",
  "reasoning": "With consistently strong performance across stages, this next arena introduces a vertical corridor structure on the left side with the target inside it, plus additional horizontal barriers in the lower-right area where the start is located, requiring the policy to navigate through a more constrained passage while maintaining path diversity from previous horizontal-dominated layouts."
}
\end{lstlisting}

\newpage
\subsection{Stage 4}

\subsubsection{\numcirc{1} Optimization}\hfill\\
\includegraphics[width=4cm]{data/good-run/imgs/s3-train.best-train.trajs_svg-raw.pdf}\\
\includegraphics[width=6cm]{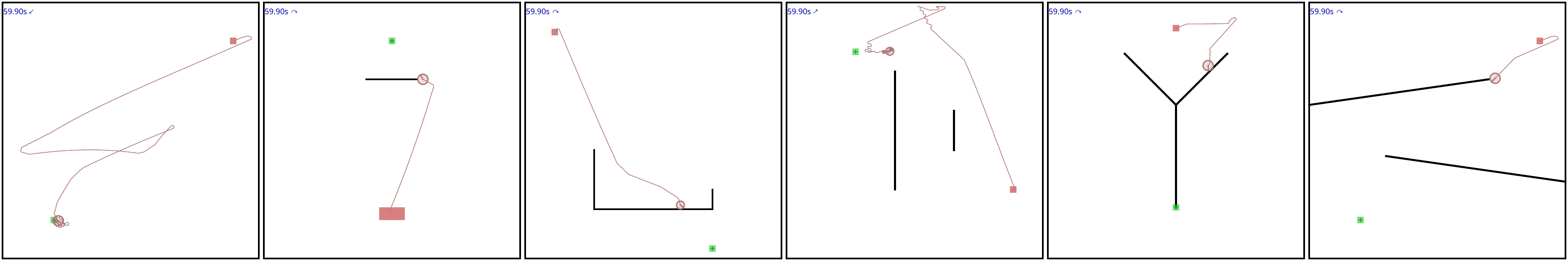}\\
\includegraphics[width=\linewidth]{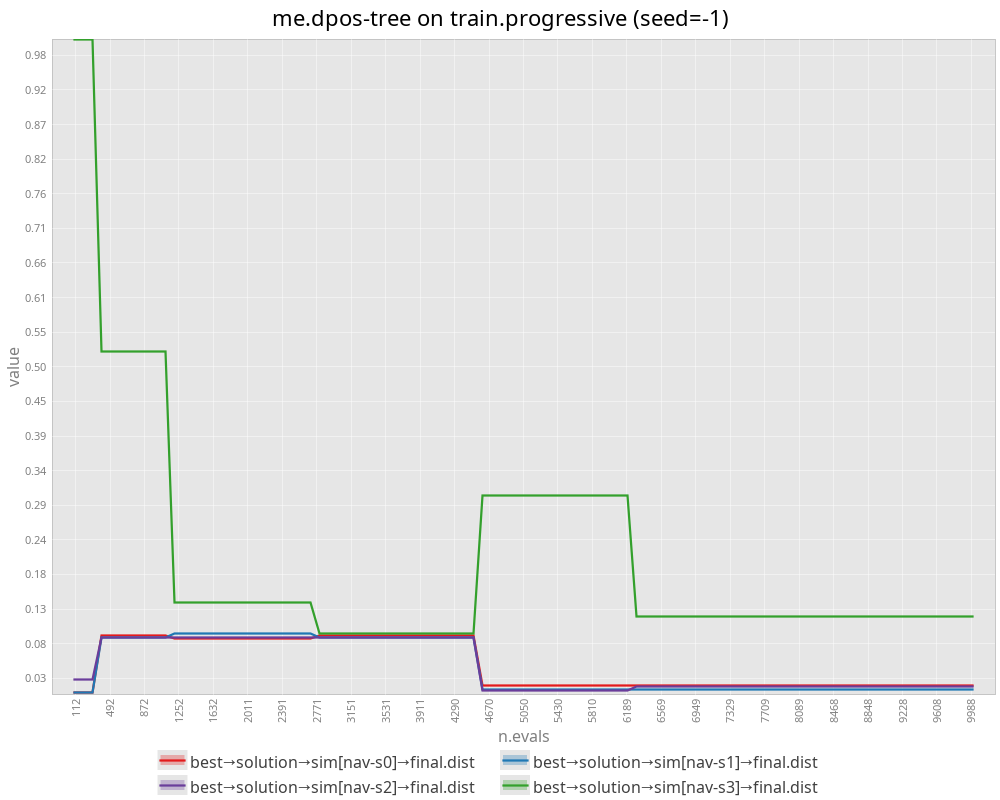}

\subsubsection{\numcirc{2} Feedback}\hfill
\begin{lstlisting}[inputencoding=utf8]
CONTRACT: STAGE-GRID v1.0 applies.
## PROVIDED FEEDBACK
- quality of the best policy on the last stage : {0.119}        
- quality of the best policy on previous stages : {0.013, 0.007, 0.013, 0.119}
- quality on average of all stages: {0.0382}
## REMINDER (do not change)
- Grid: 15x15; rows joined by '|' (do not use '|' as a cell).
- Size: exactly 15 rows × 15 columns; every row must be length 15. 
- Alphabet: {w, s, t, e}; exactly one 's' and one 't'.
- Variety: vary the positions of 's' and 't' across stages.
- Reachability: at least one 4-connected path of 'e' from 's' to 't' (no diagonals).
- Include: "understood" and "reasoning".
Don’t forget the constraints and validation check. If any output would violate CONTRACT: STAGE-GRID v1.0, regenerate internally and only emit a valid JSON object.
- the two plots are attached. The two plots are for reference only; do not reproduce them in the output.
## ROW LENGTH VALIDATION (CRITICAL)
1. Each row = exactly 15 characters
2. Total of 15 rows
3. Validation method: COUNT each row before output
4. If count ≠ 15 → REJECT and regenerate entire grid
\end{lstlisting}

\subsubsection{\numcirc{3} Response}\hfill
\begin{lstlisting}[inputencoding=utf8]
{
  "grid": "eeeeeeeeeeeeeee|eeeewwwweeeeeee|eeeeeeeeeeeeeee|eeeeeeeeeeeeeee|teeeeeeeeeeeeee|eeeeeeeewwwwwwe|eeeeeeeeeeeeeee|wwwwweeeeeeeeee|eeeeeeeeeeeeeee|eeeeeeeeeweeeee|eeeeeeeeeweeeee|eeeeeeeeewwwwwe|eeeeeeeeeeeeeee|eeeeeeeeeeeeeee|seeeeeeeeeeeeee",
  "understood": "Stage 3 reached quality 0.119 after starting at 1.0 and converging around evaluation 5000, showing a struggle compared to previous near-optimal stages; the trajectory reveals a curved path from lower-right start navigating leftward around the vertical corridor structure to reach the upper-middle target, indicating the constrained passage was challenging.",
  "reasoning": "Given the increased difficulty in stage 3 (quality 0.119 vs ~0.01 in prior stages), this stage consolidates learning with moderate complexity by placing start at bottom-left and target at top-left on the same vertical edge, introducing scattered horizontal wall segments at different heights that require navigating around but without tight corridor constraints, allowing the policy to refine navigation skills before escalating difficulty further."
}
\end{lstlisting}

\newpage
\subsection{Stage 5}

\subsubsection{\numcirc{1} Optimization}\hfill\\
\includegraphics[width=5cm]{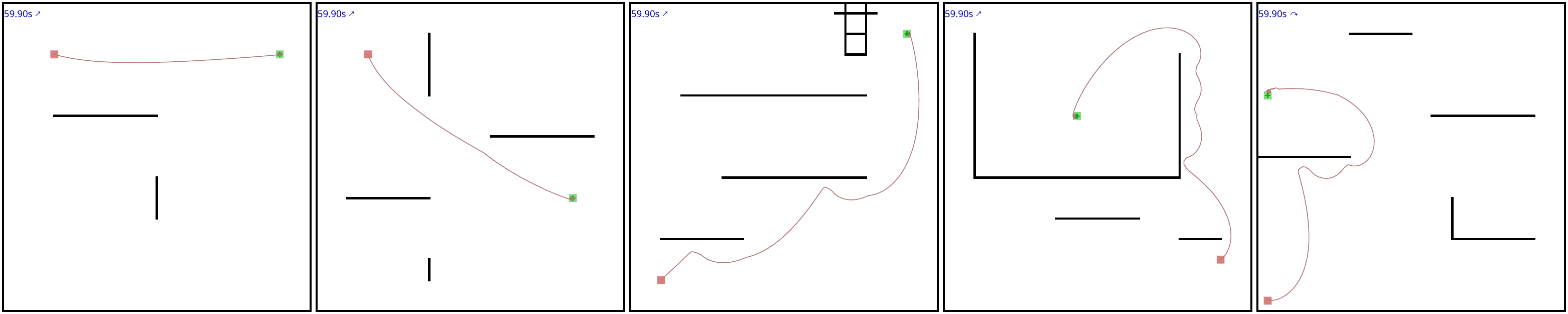}\\
\includegraphics[width=6cm]{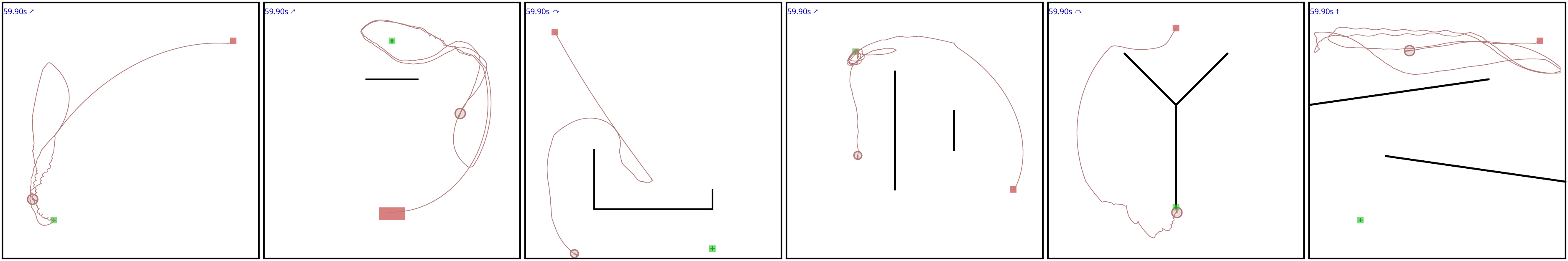}\\
\includegraphics[width=\linewidth]{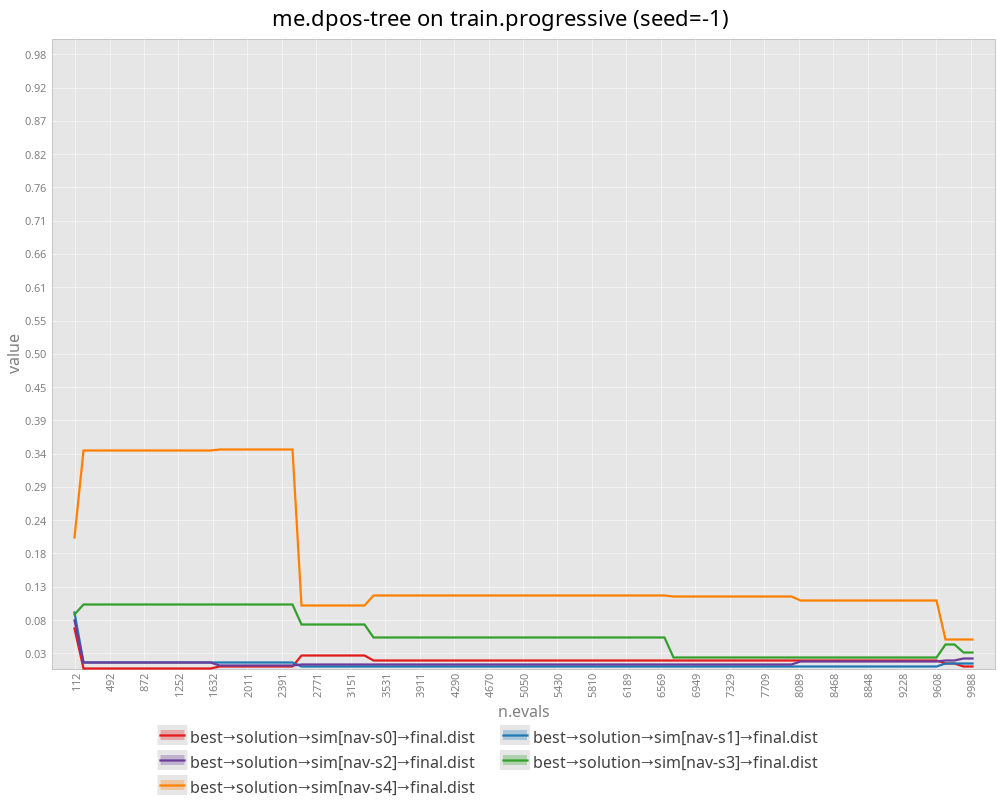}

\subsubsection{\numcirc{2} Feedback}\hfill
\begin{lstlisting}[inputencoding=utf8]
CONTRACT: STAGE-GRID v1.0 applies.
## PROVIDED FEEDBACK
- quality of the best policy on the last stage : {0.048}        
- quality of the best policy on previous stages : {0.006, 0.009, 0.017, 0.027, 0.048}
- quality on average of all stages: {0.0212}
## REMINDER (do not change)
- Grid: 15x15; rows joined by '|' (do not use '|' as a cell).
- Size: exactly 15 rows × 15 columns; every row must be length 15. 
- Alphabet: {w, s, t, e}; exactly one 's' and one 't'.
- Variety: vary the positions of 's' and 't' across stages.
- Reachability: at least one 4-connected path of 'e' from 's' to 't' (no diagonals).
- Include: "understood" and "reasoning".
Don’t forget the constraints and validation check. If any output would violate CONTRACT: STAGE-GRID v1.0, regenerate internally and only emit a valid JSON object.
- the two plots are attached. The two plots are for reference only; do not reproduce them in the output.
## ROW LENGTH VALIDATION (CRITICAL)
1. Each row = exactly 15 characters
2. Total of 15 rows
3. Validation method: COUNT each row before output
4. If count ≠ 15 → REJECT and regenerate entire grid
\end{lstlisting}

\subsubsection{\numcirc{3} Response}\hfill
\begin{lstlisting}[inputencoding=utf8]
{
  "grid": "eeeeeeeeeeeeeee|ewwwwwwweeeeeee|eeeeeeeeeeeeeee|eeeeeeeweeeeeee|eeeeeeeweeeeeee|eeteeeewwwwwwwe|eeeeeeeeeeeeeee|eeeeeeeeeeeeeee|ewwwwweeeeeeeee|eweeeeeeeeeeeee|eweeeeeeeweeeee|eweeeeeeeweeeee|eweeeeeeeweeeee|eweeeeeeewwwwwe|eeeeeeeeeeeeese",
  "understood": "Stage 4 achieved quality 0.048 after starting at 0.21 and stabilizing around evaluation 2500; the trajectory shows a complex curved path from lower-right area to upper-right target, navigating around a large rectangular enclosure and scattered horizontal barriers, demonstrating successful adaptation to constrained spatial navigation.",
  "reasoning": "With quality gradually increasing from 0.006 to 0.048 across stages, this next arena introduces a mixed wall configuration: an upper horizontal barrier, a central corridor structure combining vertical and horizontal segments, and a lower-left L-shaped corner obstacle, with start positioned in the bottom-left corner and target in the upper-left area to test diagonal navigation through multiple distinct obstacle types."
}
\end{lstlisting}

\newpage
\subsection{Stage 6}

\subsubsection{\numcirc{1} Optimization}\hfill\\
\includegraphics[width=6cm]{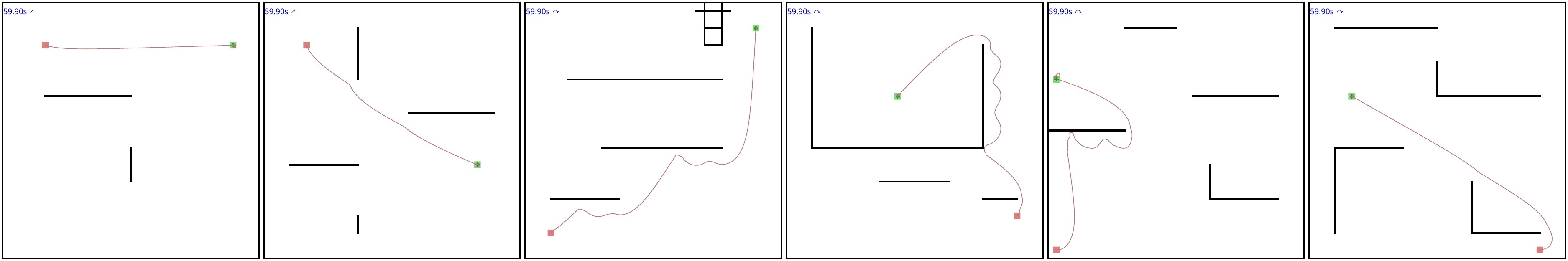}\\
\includegraphics[width=6cm]{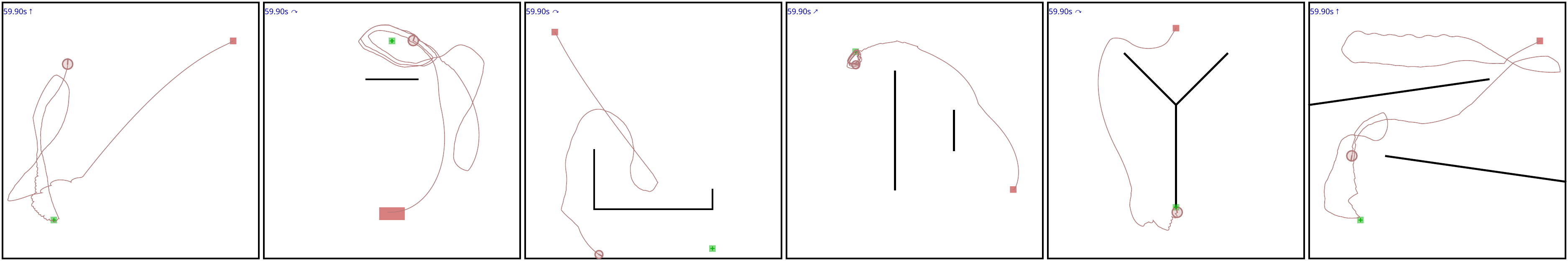}\\
\includegraphics[width=\linewidth]{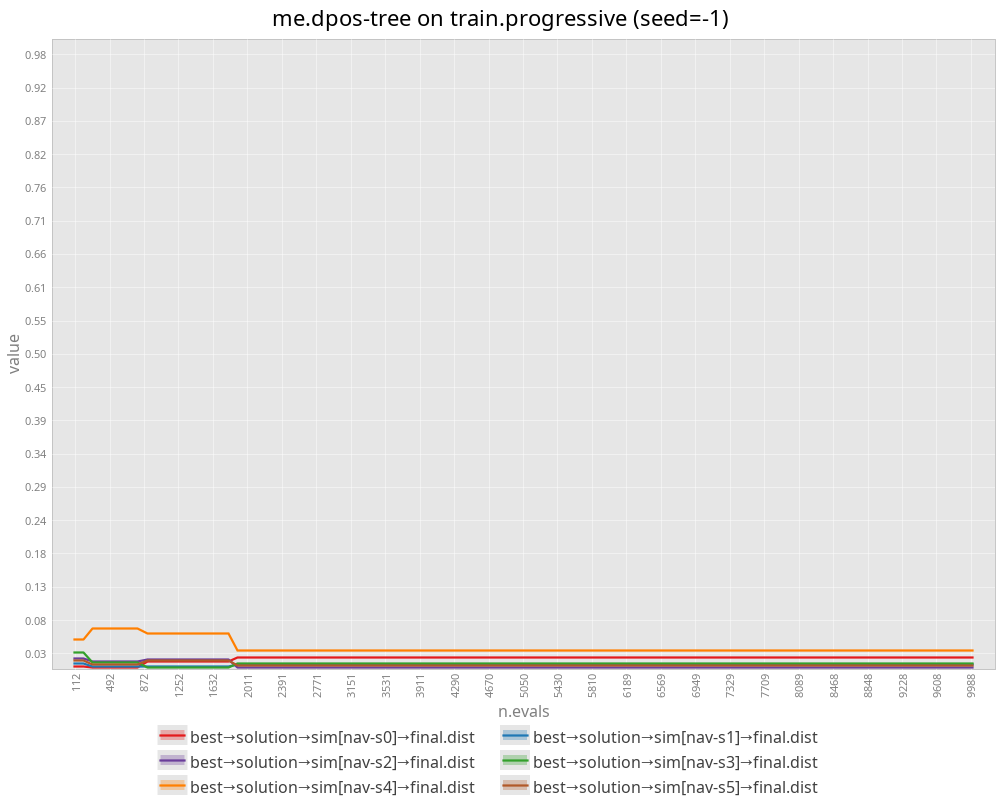}

\subsubsection{\numcirc{2} Feedback}\hfill
\begin{lstlisting}[inputencoding=utf8]
CONTRACT: STAGE-GRID v1.0 applies.
## PROVIDED FEEDBACK
- quality of the best policy on the last stage : {0.007}        
- quality of the best policy on previous stages : {0.020, 0.008, 0.003, 0.010, 0.031, 0.007}
- quality on average of all stages: {0.0129}
## REMINDER (do not change)
- Grid: 15x15; rows joined by '|' (do not use '|' as a cell).
- Size: exactly 15 rows × 15 columns; every row must be length 15. 
- Alphabet: {w, s, t, e}; exactly one 's' and one 't'.
- Variety: vary the positions of 's' and 't' across stages.
- Reachability: at least one 4-connected path of 'e' from 's' to 't' (no diagonals).
- Include: "understood" and "reasoning".
Don’t forget the constraints and validation check. If any output would violate CONTRACT: STAGE-GRID v1.0, regenerate internally and only emit a valid JSON object.
- the two plots are attached. The two plots are for reference only; do not reproduce them in the output.
## ROW LENGTH VALIDATION (CRITICAL)
1. Each row = exactly 15 characters
2. Total of 15 rows
3. Validation method: COUNT each row before output
4. If count ≠ 15 → REJECT and regenerate entire grid
\end{lstlisting}

\subsubsection{\numcirc{3} Response}\hfill
\begin{lstlisting}[inputencoding=utf8]
{
  "grid": "eeeeeeeeeeeeeee|eewwwwwwweewwwe|eeeeeeeeeeeeeee|eeeeeeeweeeeeee|wwwwweeweeeeeee|eeteeeeweeeeeee|eeeeeeeweeeeeee|eeeeeeeweeeseee|eeeeeeeweeeeeee|eeeeeeeweeeeeee|eeeeeeeeeeeewwe|eeeeeeeeeeeeeee|ewwwwwweeeeeeee|eeeeeeeeeeeeeee|eeeeeeeeeeeeeee",
  "understood": "Stage 5 achieved near-optimal quality 0.007 with rapid convergence by evaluation 1500; the trajectory shows a smooth diagonal path from upper-left to right-middle navigating around the complex multi-level corridor and L-shaped structures, demonstrating strong navigation capabilities through the varied obstacle configurations.",
  "reasoning": "With consistent strong performance (average quality 0.0129), this stage introduces a vertical corridor running through the central-right area with the start embedded within it, requiring the robot to first navigate out of the corridor and then reach the target on the left side, while upper and lower horizontal barriers create additional routing constraints to test escape-and-navigate behavior."
}
\end{lstlisting}

\newpage
\subsection{Stage 7}

\subsubsection{\numcirc{1} Optimization}\hfill\\
\includegraphics[width=7cm]{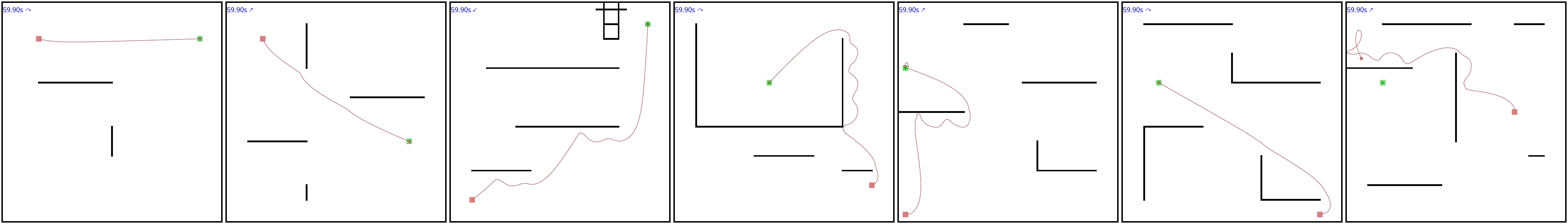}\\
\includegraphics[width=6cm]{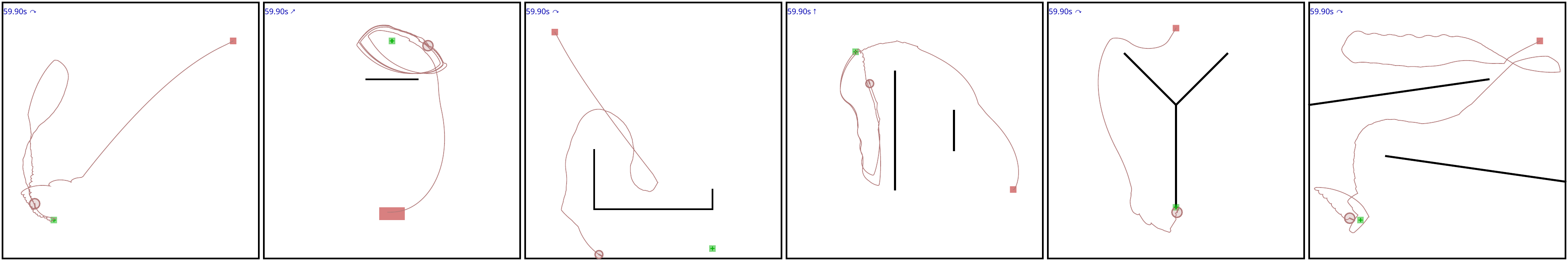}\\
\includegraphics[width=\linewidth]{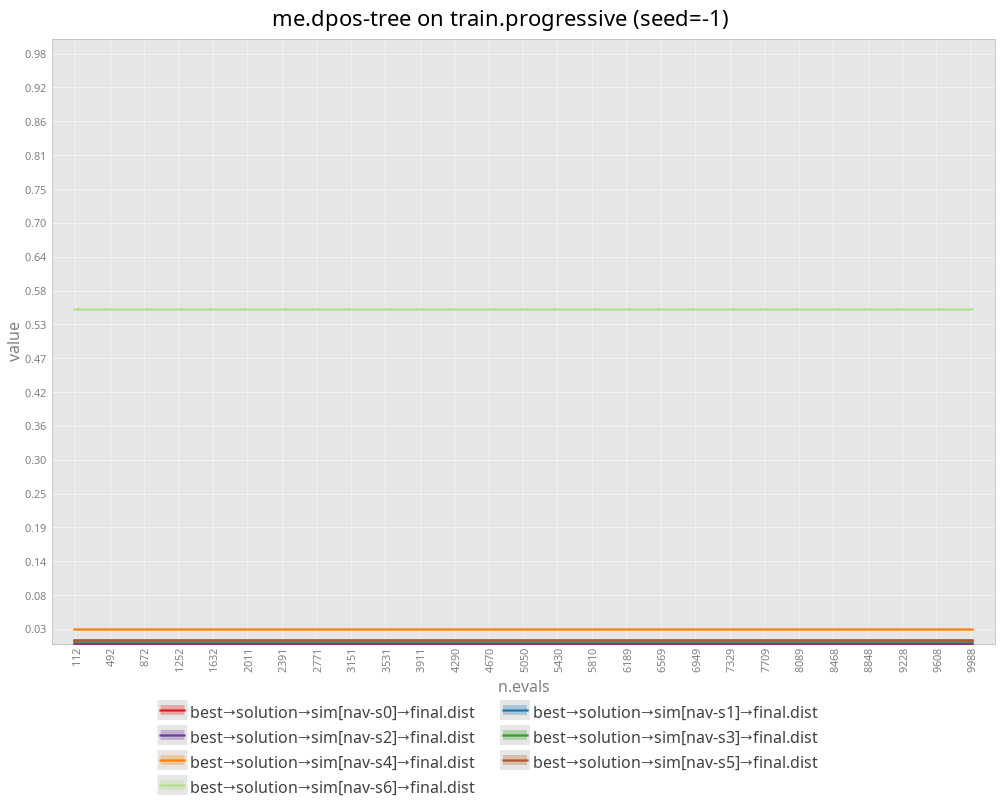}

\subsubsection{\numcirc{2} Feedback}\hfill
\begin{lstlisting}[inputencoding=utf8]
CONTRACT: STAGE-GRID v1.0 applies.
## PROVIDED FEEDBACK
- quality of the best policy on the last stage : {0.553}        
- quality of the best policy on previous stages : {0.000, 0.008, 0.002, 0.005, 0.025, 0.007, 0.553}
- quality on average of all stages: {0.0856}
## REMINDER (do not change)
- Grid: 15x15; rows joined by '|' (do not use '|' as a cell).
- Size: exactly 15 rows × 15 columns; every row must be length 15. 
- Alphabet: {w, s, t, e}; exactly one 's' and one 't'.
- Variety: vary the positions of 's' and 't' across stages.
- Reachability: at least one 4-connected path of 'e' from 's' to 't' (no diagonals).
- Include: "understood" and "reasoning".
Don’t forget the constraints and validation check. If any output would violate CONTRACT: STAGE-GRID v1.0, regenerate internally and only emit a valid JSON object.
- the two plots are attached. The two plots are for reference only; do not reproduce them in the output.
## ROW LENGTH VALIDATION (CRITICAL)
1. Each row = exactly 15 characters
2. Total of 15 rows
3. Validation method: COUNT each row before output
4. If count ≠ 15 → REJECT and regenerate entire grid
\end{lstlisting}

\subsubsection{\numcirc{3} Response}\hfill
\begin{lstlisting}[inputencoding=utf8]
{
  "grid": "eeeeeeeeeeeeeee|eeeeeeeeeeeeeee|wwwwweeeeeewwwe|eeeeeeeeeeeeeee|eeeeeeeeeeeeeee|eeeewwwwweeeeee|eeeeeeeeeeeeeee|eseeeeeeeeeteee|eeeeeeeeeeeeeee|eeeewwwwweeeeee|eeeeeeeeeeeeeee|eeeeeeeeeeeeeee|wwwwweeeeeewwwe|eeeeeeeeeeeeeee|eeeeeeeeeeeeeee",
  "understood": "Stage 6 showed quality 0.553 with a completely flat progression curve remaining at that level across all 10,000 evaluations, indicating the vertical corridor structure was too challenging; the trajectory plot reveals the robot struggled to escape the constrained passage to reach the left-side target.",
  "reasoning": "Given the significant difficulty spike in stage 6 (quality jumped from 0.007 to 0.553 with no learning progress), this stage steps back in complexity by providing an open arena with symmetrically placed horizontal wall segments at the top, middle, and bottom levels, requiring basic obstacle avoidance without tight constraints, allowing the policy to rebuild confidence and navigation skills before attempting more complex corridor-based challenges."
}
\end{lstlisting}

\subsection{Stage 8}

\subsubsection{\numcirc{1} (Last) optimization}\hfill\\
\includegraphics[width=8 cm]{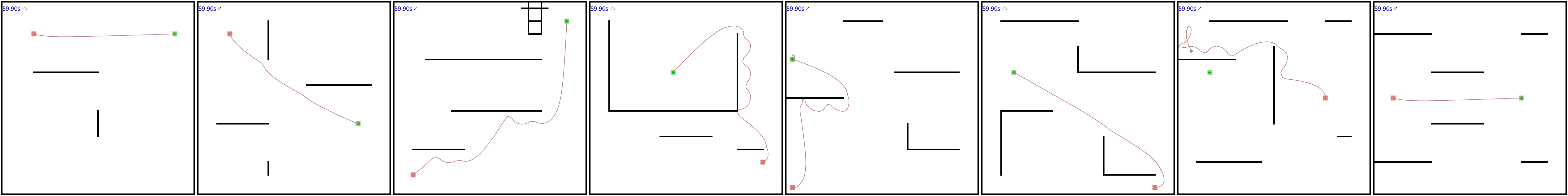}\\
\includegraphics[width=6cm]{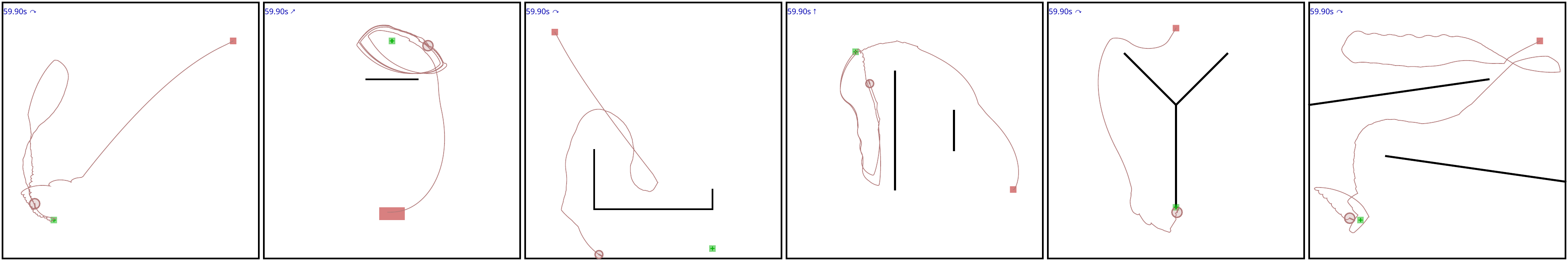}

\end{document}